\documentclass[lettersize,journal]{IEEEtran}
\usepackage[noadjust]{cite}

\usepackage{multirow,multicol}
\usepackage{slashbox}

\usepackage[tbtags]{amsmath}
\usepackage{amsbsy}
\usepackage{amssymb}
\usepackage{amsfonts}
\usepackage{bbm}

\usepackage{graphicx}
\usepackage[caption=false]{subfig}
\usepackage{url}
\usepackage{textcomp}
\usepackage{algorithmic}
\usepackage{algorithm}
\usepackage{balance}

\usepackage{array}

\usepackage{hyperref}
\usepackage{color}

\newcommand{\etal}{\textit{et al}. }
\newcommand{\ie}{\textit{i}.\textit{e}., }
\newcommand{\eg}{\textit{e}.\textit{g}., }

{\begin{list}               
    {$\bullet$ \hfill}{
        \setlength{\leftmargin}{\parindent}
        \setlength{\parsep}{0.04\baselineskip}
        \setlength{\itemsep}{0.5\parsep}
        \setlength{\labelwidth}{\leftmargin}
        \setlength{\labelsep}{0em}}
    }
{\end{list}}

\providecommand{\eref}[1]{\eqref{#1}}  
\providecommand{\cref}[1]{Chapter~\ref{#1}}
\providecommand{\sref}[1]{Section~\ref{#1}}
\providecommand{\fref}[1]{Fig.~\ref{#1}}
\providecommand{\tref}[1]{Table~\ref{#1}}

\begin{document}

\title{Sampling Agnostic Feature Representation for Long-Term Person Re-identification}

\author{Seongyeop~Yang, Byeongkeun Kang, and Yeejin Lee 
\thanks{S.~Yang and Y.~Lee are with the Department of Electrical and Information Engineering, Seoul National University of Science and Technology, Seoul, South Korea~(e-mail: syyang@seoultech.ac.kr, yeejinlee@seoultech.ac.kr).}
\thanks{B. Kang is with the Department of Electronic Engineering, Seoul National University of Science and Technology, Seoul, South Korea~(e-mail: byeongkeun.kang@seoultech.ac.kr).}
}



\maketitle

\begin{abstract}
Person re-identification is a problem of identifying individuals across non-overlapping cameras. Although remarkable progress has been made in the re-identification problem, it is still a challenging problem due to appearance variations of the same person as well as other people of similar appearance. Some prior works solved the issues by separating features of positive samples from features of negative ones. However, the performances of existing models considerably depend on the characteristics and statistics of the samples used for training. Thus, we propose a novel framework named sampling independent robust feature representation network~(SirNet) that learns disentangled feature embedding from randomly chosen samples. A carefully designed sampling independent maximum discrepancy loss is introduced to model samples of the same person as a cluster. As a result, the proposed framework can generate additional hard negatives/positives using the learned features, which results in better discriminability from other identities. Extensive experimental results on large-scale benchmark datasets verify that the proposed model is more effective than prior state-of-the-art models.  
\end{abstract}

\begin{IEEEkeywords}
Long-term person re-identification, data mining, feature augmentation, classification loss
\end{IEEEkeywords}

\section{Introduction} \label{sec:intro}
\IEEEPARstart{P}{erson} re-identification~(ReID) is an essential task to discover the identity of individuals across multiple non-overlapping cameras. It has been a major topic of research in computer vision for the past few decades due to the significant role it plays in various applications~\cite{zheng2016person, ye2021deep, 8294254}, including intelligent surveillance systems, security systems,  self-driving vehicles, and computational forensics.

When a person is captured by multiple cameras, recent re-identification models usually identify a person using the features learned from appearance cues, for example, color, contrast, texture, and clothing style~\cite{zheng2019joint, zheng2015scalable, Li_2018_CVPR, Zhao_2017_CVPR}. However, in a more realistic real-world re-identification task, there could be many people in public spaces having similar-looking appearances, making it hard to differentiate identities. Meanwhile, when a person is captured by different cameras across several days, his/her appearance can be significantly changed. Due to such appearance changes and environmental uncertainty, re-identification modeling is much more challenging in the long-term scenarios~\cite{yu2019robust, ye2021deep, Huang_2020_TCSVT, hong2021fine}, although remarkable progress has been made in the short-term scenarios~\cite{zheng2019joint, Sun_2020_CVPR, Jin_2020_CVPR, Lin_2020_CVPR}.

\begin{figure}[!t]
    \begin{minipage}{0.40\linewidth}
        \hspace{0.5cm}
        \centerline{\includegraphics[scale=0.31]{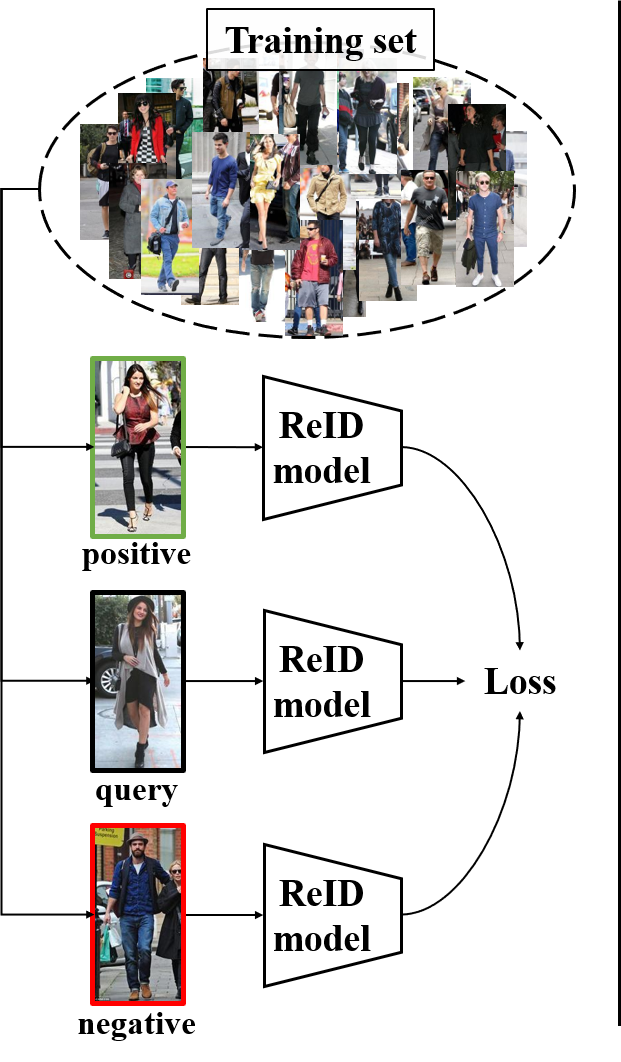}}
    \end{minipage}
    \begin{minipage}{0.50\linewidth}
        \hspace{1.7cm}
        \centerline{\includegraphics[scale=0.31]{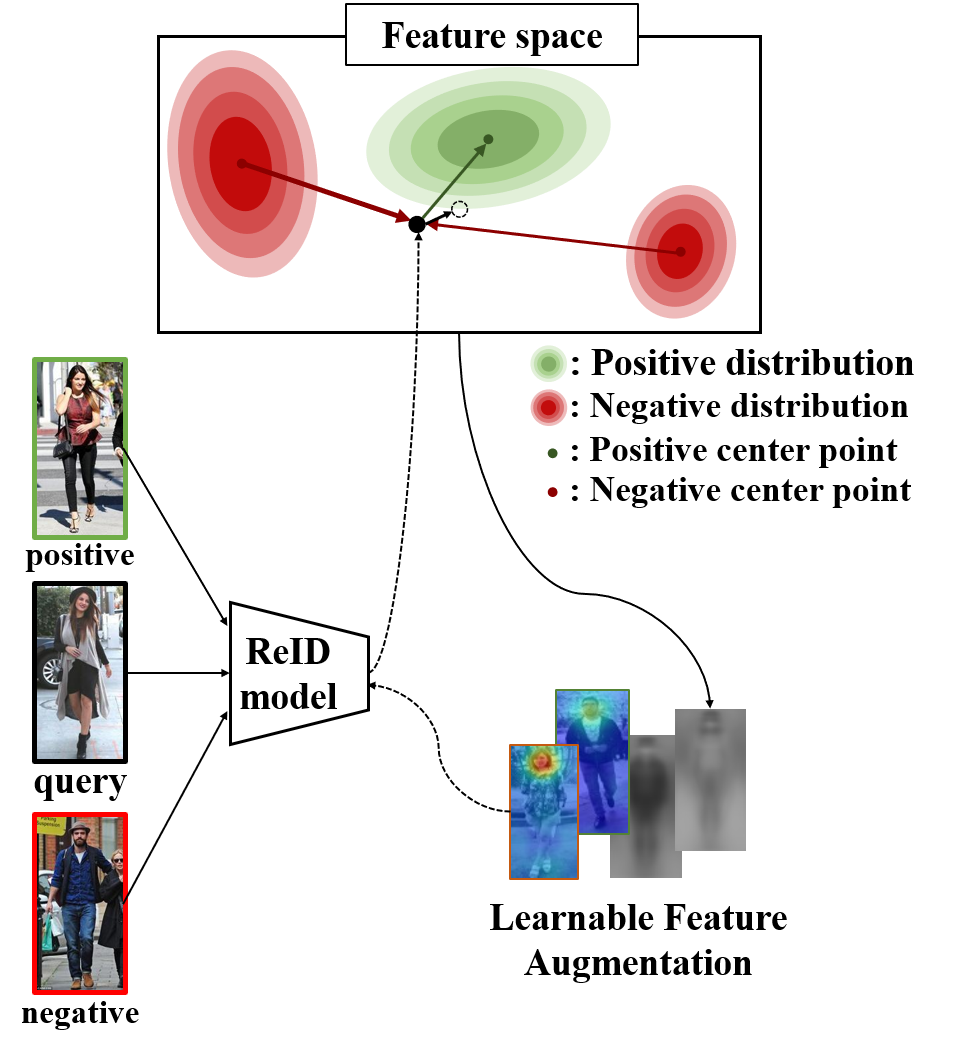}}
    \end{minipage}
    \caption{How to select samples for training? The proposed framework learns feature representation measuring the distances of clusters in the feature space together with conventional similarity measures, as described in the figure on the right side. By the learned feature representation, the proposed framework augments difficult samples as the model trains. This ensures that the proposed framework enhances intra-identity indiscrimination and inter-identity discriminability. In contrast, the performance of a model treating each sample as a feature vector is considerably dependent on the selected samples in the figure on the left side.}  \label{fig:concept}
\end{figure}

In order to overcome appearance variation in long-term scenarios, most recent re-identification models try to learn deep feature representations by separating them into id-relevant and id-irrelevant embeddings. Several works use pose estimation to detect body parts~\cite{huang2019celebrities, wan2020person,yu2020cocas,qian2020long}, which learn part-specific features and thus eliminate the features possibly related to appearance. Similarly, some works take the results of contour estimation~\cite{yang2019person} or human mask extraction~\cite{hong2021fine} as inputs on which the models learn body shape features. More recently, the works~\cite{xuadversarial, li2020learning} use adversarial learning to decouple features from colors and features from body shapes. Similar to them, to separate features, recent approaches learn id-irrelevant features from inputs of different people~(negatives) and id-relevant features from inputs of the same person~(positives) guided by similarity learning objectives, such as triplet loss~\cite{schroff2015facenet}. 

Although triplet-based distance metric is commonly used for feature disentanglement, the performances of models using it are considerably dependent on which samples to be used as inputs. Different sampling strategies lead to drastically different solutions even for the same loss function~\cite{schroff2015facenet, wu2017sampling, zhai2019defense}. This performance variation can be alleviated by selecting non-easily trainable samples~\cite{schroff2015facenet, wu2017sampling, zhai2020ad, chen2020hard, shu2021large}. Specifically, a model cannot learn meaningful features from different people with distinct appearances. On the other hand, the same person in different appearances~(hard positive) enforces the model to maximizing intra-class variance. And similar-looking but different people~(hard negative) allow the model to enhance inter-class discriminability. However, there are difficulties in mining such ``good'' samples. Since computing selection metrics across a whole training set is infeasible in large-scale datasets, conventional models usually choose samples within a mini-batch~\cite{schroff2015facenet, hermans2017defense, shu2021large}. In addition, hard sample selection might lead to poor training performance because poor quality data, such as mislabelled or noisy, would dominate in hard samples~\cite{schroff2015facenet,yu2019robust}. Even worse, selecting too hard samples can make the training unstable~\cite{hermans2017defense}. Furthermore, grouping the samples of a person with different visual styles can sacrifice inter-class discriminability. That is, increasing identity-specific clusters can intrude clusters of other identities, although maximizing intra-class variance is essential for long-term re-identification tasks with significant personal style changes. These two complementary objectives, increasing compactness of identity clusters while reducing overlap, make long-term re-identification harder than short-term re-identification.

To overcome the aforementioned data sampling difficulties, in this paper, we propose a framework that learns feature embedding from randomly chosen samples without applying any hard sample mining strategy. In addition, the proposed framework represents each sample as a cluster of points belonging to the same person to tackle the intra-identity variation issue. In the proposed framework, distinct identity feature embedding is learned by measuring the distances of cluster centers, regularly interacting with all samples. This differs from conventional models representing each sample as a feature vector disjointing with feature points corresponding to the same person. As a consequence of the disentangled feature representation, the proposed framework can generate hard samples using the learned features embeddings, making the model well-generalizable on problem sets. To do this, id-relevant and id-irrelevant features from positive and negative samples are re-entangled using a class activation map. This learnable feature augmentation ensures consistently increasing difficulty of samples as the model trains. And owing to this learned feature augmentation, the proposed model is less prone to be attacked by images of different identities but similar appearances and images of a person with various appearance changes.

In summary, the contributions of this work are as follows: (1) We propose a novel framework named \underline{S}ampling \underline{I}ndependent \underline{R}obust Feature Representation Network~(\textit{SirNet}). The proposed \textit{SirNet} models feature points as a cluster that is robust against appearance changes as well as discriminative against different identities. (2) We present a learnable feature augmentation method to generate hard samples, which improves the discriminability of identities and generalization capability on unseen datasets. (3) Extensive comparative evaluations demonstrate the effectiveness of the proposed framework against the current state-of-art models on four benchmark datasets.

The remainder of the paper is organized as follows. \sref{sec:related_work} reviews the state-of-the-art long-term re-identification models and data mining strategies for re-identification. \sref{sec:proposed} describes the framework of \textit{SirNet} proposed for learning robust feature representation in long-term person re-identification tasks. \sref{sec:results} presents the experiments and discusses the experimental results, and concluding remarks are made in \sref{sec:conclusion}.  

\section{Related Works} \label{sec:related_work}
\subsection{Long-Term Person ReID}
The recent publication of large-scale long-term person re-identification datasets has enabled the models to benefit from more advanced supervised learning-based techniques. Early work by Lee \etal built a wardrobe model that captured each person's set of clothes. This work used clothing as a soft biometric for re-id and tracking where clothing items were assumed to be constant~\cite{lee2018wardrobe}.

As opposed to retrieving clothing consistency, recent works address the problem of appearance-oriented descriptive labeling in long-term re-identification tasks. 
Huang \etal proposed a two-step fine-tuning approach using full images and body part images~(2SF-BPart) based on the 2S-IDE network~\cite{huang2019celebrities}. They later designed a ReIDCaps network to learn id-related features adopting soft embedding attention and feature sparse representation~\cite{Huang_2020_TCSVT}. In the work, the proposed model utilized vector neurons that helped to recognize the same person under clothe-changing. 
Yang \etal used a deep neural network to transform contour sketch images for extracting features in polar coordinate space to overcome moderate clothe changing~\cite{yang2019person}. Wan \etal proposed a 3-stream appearance, part, and face extractor network~(3APF) that utilized pose estimation and face detection~\cite{wan2020person}. Yu \etal proposed a biometric-clothes network~(BC-Net), aided by a clothes detector and learnable masks to focus on clothing and body regions~\cite{yu2020cocas}. Qian \etal used full images and body keypoints to disentangle clothe features and body shape features using cloth-elimination shape-distillation module~(CESD) to eliminate clothes information and focus on the body shape information~\cite{qian2020long}. Hong \etal explored a fine-grained shape-appearance mutual learning framework~(FSAM) that learned human masks and soft pose labeling~\cite{hong2021fine}. However, most of the aforementioned methods only can eliminate certain identity-irrelevant features, such as pose, clothes, or pose/clothes, not considering other variations in long-term re-identification. Moreover, they require defining what kinds of id-irrelevant information are used for feature disentanglement.

Similar to the work~\cite{zheng2019joint}, a generative adversarial network~(GAN) is employed for long-term re-identification~\cite{xuadversarial, li2020learning}. Xu \etal proposed an adversarial feature disentanglement network~(AFD-Net) that mapped a person's image into appearance/structure and reassembled them to new person images~\cite{xuadversarial}. Li \etal proposed a model, color agnostic shape extraction network~(CASE-Net), which took a gray-scale image as an input to remove color variations~\cite{li2020learning}. Recently, Huang \etal proposed the clothing status awareness network~(RCSANet) that handles clothes changing by introducing an alternative weighted triplet loss over a mini-batch~\cite{huang2021clothing}.

Note that the learning objectives of the proposed framework allow joint comparisons of the feature points of all negative samples as well as  the feature points of all positives across the entire dataset, while the previously mentioned works employ only a few negative points and positive points over a mini-batch. The proposed framework can generate augmented samples using the disentangled features, similar to generative adversarial models. However, unlike the works in \cite{xuadversarial} and \cite{li2020learning}, the proposed framework does not require the generative adversarial model, which is tricky to train. Moreover, the proposed framework adopts feature-level augmentation and does not require additional networks for regenerating images from features; thus, its architecture is more compact than the work in \cite{xuadversarial}. In addition to these structural and training advancements, the proposed framework disentangles features without any limited configuration on input pairs. While the works in \cite{xuadversarial} and \cite{li2020learning} take input pairs containing the same person images with different poses~(the input pairs are necessary to be carefully selected), the inputs of the proposed framework are randomly selected. This also means that the proposed framework only takes RGB images without any other cues~(\eg mask, segmentation, depth) that require additional computations.

\subsection{Sample Mining for ReID}
Although it is well accepted that the performance of deep metric loss is significantly affected by sampling methods~\cite{wu2017sampling}, the problem of data selection has been less studied in person re-identification. Schroff \etal used semi-hard negative exemplars that were further away from the anchor than the positive exemplar, but
still hard because the squared distance was close to the positive distance~\cite{schroff2015facenet}. Hermans \etal proposed the online variants of offline hard negative mining~\cite{hermans2017defense}. They empirically proved that batch hard loss with the soft-margin variation worked the best. Yu \etal proposed a soft hard sample mining scheme by adaptively assigning weights to hard samples~\cite{yu2018hard}. Recently, Shu \etal selected positive sample pairs of the maximum distance to an anchor and negative sample pairs of the minimum distance within a mini-batch~\cite{shu2021large}, similar to the works in \cite{schroff2015facenet,hermans2017defense}. Zhang \etal proposed a framework that simultaneously trained a model with selected samples and the data sampler~\cite{zhang2021one}. 

Unlike the previously mentioned works, the proposed framework uses randomly selected samples without any hard sample mining. Nevertheless, it can maximize inter-class variations and intra-class diversity by introducing a loss function that measures the distance of identity clusters over all samples. 

\section{Proposed Framework} \label{sec:proposed}


In this section, we provide the details of the proposed {\textit{SirNet}} framework for long-term person re-identification. We address the problem in \sref{sec:problem_definition} and develop main components of the proposed framework in \sref{sec:framework_overview}. \sref{sec:identity_loss} derives the sample independent loss over feature distribution, and \sref{sec:reconstruction_loss} describes feature augmentations to generate hard samples during training.

\subsection{Preliminaries} 
\label{sec:problem_definition}
\begin{figure*}[!t]
    \centering
    \includegraphics[width=0.75\textwidth]{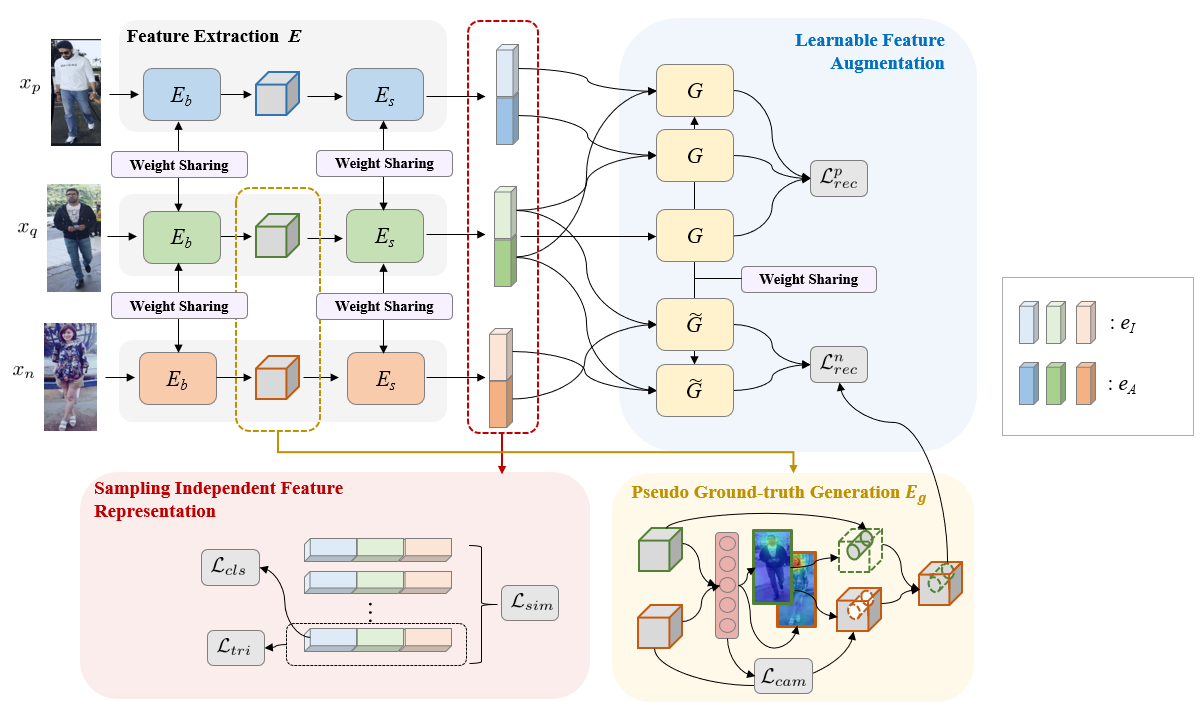} 
    \caption{The overall architecture of the proposed framework. Given a triplet, the feature extraction module $E$ maps $x$ into id-relevant and id-irrelevant embeddings. With the learned embeddings, the feature augmentation module $G$ generates hard positive and hard negative samples emanated from the pseudo-ground-truth generation module $E_g$. The proposed framework models a single sample by a cluster, maximizing intra-class diversity and inter-class discriminability.} \label{fig:overall_architecture}
\end{figure*}
Given a set of $N_g$ gallery images and its corresponding label pairs $\mathcal{G} =\left\{\left(x_{\scriptscriptstyle g}, y_{\scriptscriptstyle g}\right) \:\rvert\: g = 1, 2, \cdots, N_g\right\}$, a person re-identification problem can be formulated as selecting discrete labels of the relevant images for a query image $x_{\scriptscriptstyle q}$ among a large number of candidate images. Specifically, a set of $k$ labels $\mathcal{R}(x_{\scriptscriptstyle q}, k)$ determines which images are the closest to $x_{\scriptscriptstyle q}$ in embedding space:
\begin{align}
\mathcal{R}(x_{\scriptscriptstyle q}, k) = \left\{ y_{\scriptscriptstyle g} \:\rvert\:  y_{\scriptscriptstyle g} \in \mathcal{N}(x_{\scriptscriptstyle q},k) \right\},
\end{align}
where $\mathcal{N}(x_{\scriptscriptstyle q},k)$ denotes 
the set of the labels of the $k$ most similar samples from $x_{\scriptscriptstyle q}$. By denoting the embedding as $E(x)$ from an input image $x$ into a feature space $\mathbb{R}^d$, the goal of a long-term person re-identification problem can be thought of as finding a mapping that represents a person's identity, independent of appearance, pose, and other variational conditions even from others of similar appearance, as addressed in \sref{sec:intro}.  

\subsection{Framework Representation} 
\label{sec:framework_overview}
Here, we propose a framework that models a disentangled feature representation and thus can generate hard augmented samples itself. The proposed framework tackles intra-class~(person) variance and inter-class invariance problems by introducing sample independent loss on classification and reconstruction loss on augmented features. The framework we propose is illustrated in \fref{fig:overall_architecture}. The main components of the proposed framework are:
\begin{itemize}
    \item \textit{Feature Extraction} module, $E(\cdot)$ consists of weight-sharing~\cite{hoffer2015deep} backbone network $E_{b}$ followed by a weight-sharing feature separation network $E_{s}$, mapping $x$ into a representation vector $e = E(x) = E_{s}\left( E_{b}(x)\right) \in \mathbb{R}^{d}$. The module $E$ takes triplet samples and transforms them to feature representation vectors decomposed into id-relevant embedding $e_{\scriptscriptstyle I} \in \mathbb{R}^{d_{\scriptscriptstyle I}}$ and id-irrelevant embedding $e_{\scriptscriptstyle A} \in \mathbb{R}^{d_{\scriptscriptstyle A}}$, where the sum of  $d_{\scriptscriptstyle I} $ and $d_{\scriptscriptstyle A}$ is $d$.
    \item \textit{Feature Augmentation} module, weight-sharing $G(\cdot)$ generates four augmentations that contain a subset of features in a query and a subset in positive/negative samples. Additionally, two augmentations originating from $E_{g}$ generate a new set of pseudo-ground-truth. These augmentations are the re-entanglement of id-relevant features of the query with id-irrelevant features of negatives, vice versa, to produce hard samples. \sref{sec:reconstruction_loss} gives details of the augmentations. 
    \item \textit{Sampling Independent Feature Representation} module minimizes the distance of each point belonging to the same person to its center $c$ in the embedding space. At the same time, the module maximizes center distances over  all other negative identities, thus improving the model's discriminability. The details of this feature representation are in \sref{sec:identity_loss}. \\
\end{itemize}

\noindent In the proposed framework, we strive for the feature embedding $E$ by maximizing the inter-class variance and simultaneously increasing inter-class separability. For this purpose, we train the proposed framework by minimizing the below objective function
\begin{align}\label{eq:total_loss_1}
    \min\limits_{E,G, E_g} \:\:\mathcal{L}_{total},
\end{align}
where $\mathcal{L}_{total}$ is the total loss function. The total loss function is defined as a weighted sum of the identity loss $\mathcal{L}_{id}$ and the reconstruction loss $\mathcal{L}_{rec}$:
\begin{align} \label{eq:total_loss_2}
    \mathcal{L}_{total} = \lambda_{id}\mathcal{L}_{id}\:+\:\lambda_{rec}\mathcal{L}_{rec}, 
\end{align}
where $\lambda_{id}$ and $\lambda_{rec}$ are the weighting factor for each loss function. 
The details of the identity loss function and the reconstruction loss function are derived in the following \sref{sec:identity_loss} and \sref{sec:reconstruction_loss}.

In the inference stage, the concatenated output of $E(x)$ and $\alpha E_b(x)$ is used for similarity distance calculation, where $\alpha$ determines the contribution on similarity measure. 
   
\subsection{Sampling Independent Feature Embedding}
\label{sec:identity_loss}
We compute the identity loss as the combination of the cross-entropy loss and the triplet loss as following common practice~\cite{schroff2015facenet, Luo_2019_CVPR_Workshops, Cheng_2016_CVPR, 7185403, li2020learning, yang2019person, yu2020cocas,shu2021large,hermans2017defense,yu2018hard}. In addition to these conventional loss functions, we introduce a new loss term named sampling independent maximum discrepancy loss.

When a query sample $x_{\scriptscriptstyle q}$ is fed into $E$, $x_{\scriptscriptstyle q}$ maps to $e_{\scriptscriptstyle I,q}$ and $e_{\scriptscriptstyle A,q}$. Similarly, the samples $x_{\scriptscriptstyle p}$ and $x_{\scriptscriptstyle n}$ map to the feature representations $e_{\scriptscriptstyle t, p}$ and $e_{\scriptscriptstyle t, n}$, where the index $t \in \{I, A\}$ denotes id-relevant and id-irrelevant embeddings; and the subscripts $p$ and $n$ represent the positive~($y_{\scriptscriptstyle q} = y_{\scriptscriptstyle p}$) and negative~($y_{\scriptscriptstyle q} \neq y_{\scriptscriptstyle n}$) pairs of the query sample $x_{\scriptscriptstyle q}$.
In the person re-identification, the triplet loss plays a role in that the distance of the same identity is small, whereas the distance between images from different identities is large. Given a query sample, the triplet loss tries to keep embeddings of samples belonging to the same person closer than embeddings of any belonging to other people:    
\begin{align}\label{eq:L_tri}
     \mathcal{L}_{tri}\!=\! \frac{1}{N_b}\!\sum\limits_{j}^{N_b} \max \left[D_{\scriptscriptstyle T}(e_{\scriptscriptstyle I, q, j}, e_{\scriptscriptstyle I, p, j})
     \!-\!D_{\scriptscriptstyle T}(e_{\scriptscriptstyle I, q, j}, e_{\scriptscriptstyle I, n, j}) + m, 0 \right],
\end{align} 
where $D_{T}(e_{\scriptscriptstyle i}, e_{\scriptscriptstyle i^\prime}): \mathbb{R}^{\scriptscriptstyle d_{I}} \times \mathbb{R}^{\scriptscriptstyle d_{I}} \mapsto \mathbb{R}$ is a metric function that measures distances in the embedding space; $N_b$ is the number of images in a mini-batch, and $j$ indexes triplets in a mini-batch. We define the distance $D_{\scriptscriptstyle T}(\cdot)$ as the squared Euclidean norm between two id-relevant feature vectors, $D_{\scriptscriptstyle T}(e_{\scriptscriptstyle i, j}, e_{\scriptscriptstyle i^\prime, j}) :=  \left(e_{\scriptscriptstyle I, i, j} - e_{\scriptscriptstyle I, i^\prime, j}\right)^{T}\left(e_{\scriptscriptstyle I, i, j} - e_{\scriptscriptstyle I, i^\prime, j}\right)$. 

The loss \eref{eq:L_tri} ensures that given a query point $e_{\scriptscriptstyle I, q}$, the embedding of a positive point $e_{\scriptscriptstyle I, p}$ is closer to the embedding of a negative point $e_{\scriptscriptstyle I, n}$ by at least a margin $m$ for each sample. If the loss is optimized over the entire training set, all possible triplets are examined, and the points from the same person eventually pull together. However, as datasets get large, examining all possible triplets is infeasible, thus the model can converge quickly with trivial samples within a mini-batch. Even worse, the model cannot learn normal associations from non-properly selected outlier samples~\cite{hermans2017defense}.      
This implies that it is crucial to select good triplets, which contribute to training as the model's performance is significantly affected by the feature vectors of the selected set in \eref{eq:L_tri}. Specifically, to maximize the model's performance, given a query image, we want to select a sample of the same person visually dissimilar~(\ie $\mathrm{arg}\max_{p} D_{\scriptscriptstyle T}(e_{\scriptscriptstyle I, q}, e_{\scriptscriptstyle I, p})$) to learn generic features. On the other hand, to learn robust discriminative features, we want to select a sample of any other people that resemble the query person such that $\mathrm{arg}\min_{n} D_{\scriptscriptstyle T}(e_{\scriptscriptstyle I, q}, e_{\scriptscriptstyle I, n})$. 
This can be done by selecting hard positive/negative samples~\cite{schroff2015facenet, shu2021large, chen2020hard}, which explains feature representation by a specific pair only. 

To compensate for such sample-dependent feature representation, we add the sampling independent maximum discrepancy loss. The loss encourages the encoder to give closely aligned feature representation over all samples from the same person and discriminability from any other people. Concretely, the loss enforces that clusters of points belonging to the same person are pulled together in the feature embedding space while simultaneously pushing apart clusters from different persons, allowing intra-identity variations~(See \fref{fig:tsne_cluster}). 
Suppose there are $N_c$ classes and the $i$-th class has $N_i$ samples. The center $c_{\scriptscriptstyle i}$ of the $i$-th identity in feature embedding space is calculated as $c_{\scriptscriptstyle i} = 1/N_{i} \sum_{\substack{{i^\prime=1, y_{\scriptscriptstyle i^\prime}=i}}}^{N_i}\: e_{\scriptscriptstyle I, i^\prime}$, and $D_{\scriptscriptstyle c}\left(e_{\scriptscriptstyle I, j}, c_{\scriptscriptstyle i} \right): \mathbb{R}^{\scriptscriptstyle d_I} \times \mathbb{R}^{\scriptscriptstyle d_I} \mapsto \mathbb{R}$ measures the distance of a point to a cluster center, $D_{\scriptscriptstyle c}\left(e_{\scriptscriptstyle I, j}, c_{\scriptscriptstyle i} \right) = \left(e_{\scriptscriptstyle I, j} - c_{\scriptscriptstyle i}\right)^{T}\left(e_{\scriptscriptstyle I, j} - c_{\scriptscriptstyle i}\right)$. The sample independent maximum discrepancy loss $\mathcal{L}_{sim}$ tries to minimize the distance of a point to the cluster of a positive class $q$ and maximize to the clusters of all negative classes, as follows:
\begin{align}\label{eq:L_sim}
    \mathcal{L}_{sim} = 
    -\frac{1}{N_b}\sum\limits_{j}^{N_b} \: \log{ \:\: \frac{\exp{\left(-D_{\scriptscriptstyle c}(e_{\scriptscriptstyle I, q, j}, c_{\scriptscriptstyle q})\right)}}{\sum_{i}^{N_{c}}\exp{\left(-D_{\scriptscriptstyle c}(e_{\scriptscriptstyle I, q, j}, c_{\scriptscriptstyle i})\right)}}}, 
\end{align}
where the components of the embedding $e$ are normalized to less than $1$, and $i$ indexes identities. The loss is designed to be beneficial to pull together samples from the same class as much as possible while pushing samples from different classes as far as possible, as demonstrated in the results of \fref{fig:tsne_cluster}, \fref{fig:ranking}, and \tref{tab:ablation_loss}.

Recalling \eref{eq:L_tri} and \eref{eq:L_sim}, the total identity loss with the weighting factors $\lambda_{cls}$, $\lambda_{tri}$, and $\lambda_{sim}$ is defined as follows:
\begin{align}\label{eq:id_loss}
    \mathcal{L}_{id} = \lambda_{cls}\mathcal{L}_{cls} \:+\:\lambda_{tri}\mathcal{L}_{tri}\:+\:\lambda_{sim}\mathcal{L}_{sim}, 
\end{align}
and the cross-entropy loss is computed as  
\begin{align}\label{eq:cls_loss_feature}
    \mathcal{L}_{cls} = -\frac{1}{N_b}\sum\limits_{j}^{N_b} y_{\scriptscriptstyle q, j} \log\left( p\left( y_{\scriptscriptstyle q, j} | E(x_{\scriptscriptstyle q, j}) \right)\right), 
\end{align}
where $p(\cdot)$ is the predicted probability of sample $x_{\scriptscriptstyle j}$ belonging to the identity $y_{\scriptscriptstyle j}$.

In practice, for training efficiency, we train the proposed framework using triplets in conjunction with regularly updated identity clusters in \eref{eq:L_sim}. In each learning iteration, triplet loss is computed using randomly chosen from a whole training set. Alternatively, identity clusters are then updated every few epochs, using the most recent network checkpoint and computing the centers of identities for the entire training set.  
\subsection{Learnable Feature Augmentation} 
\label{sec:reconstruction_loss}
As the proposed framework implicitly learns the distribution of each identity using disentangled feature embeddings through the generative module,  it can generate a new set of hard samples to improve discriminability. This is achieved by two types of augmentations using a subset of learned features from a query and a subset of learned features from either a positive or negative sample.  
\subsubsection{Positive Feature Augmentation} For positive samples, two kinds of augmentations are considered to increase the diversity of positives. Given a subset $\left( x_{\scriptscriptstyle q}, \: x_{\scriptscriptstyle p}\right)$ of triplet, the encoder module produces feature embeddings: id-relevant embeddings $e_{\scriptscriptstyle I, q} \:, \: e_{\scriptscriptstyle I, p}$, and id-irrelevant embeddings  $e_{\scriptscriptstyle A, q} \: , \: e_{\scriptscriptstyle A, p}$. 
With positive samples, although embeddings $e_{\scriptscriptstyle A, q}$ and $e_{\scriptscriptstyle A, p}$ can generate images of different identities assigned, we want the model to produce embeddings $e_{\scriptscriptstyle I, q}$ and $e_{\scriptscriptstyle I, p}$ that can reconstruct images containing distinct id-relevant features.
To deal with this, two augmentation are considered, where the same identity of a query image is assigned to the augmented samples, allowing intra-class variations. 
We feed augmented input sets and the reconstructed input to the generative module $G$: two augmentations $\left\{ e_{\scriptscriptstyle I, p}, \:e_{\scriptscriptstyle A, q} \right\}$, $\left\{ e_{\scriptscriptstyle I, q}, \:e_{\scriptscriptstyle A, p} \right\}$, and the reconstruction $\left\{ e_{\scriptscriptstyle I, q}, \:e_{\scriptscriptstyle A, q} \right\}$, which all correspond to a person of the query image. The reconstruction loss of these positive feature augmentations measures the similarity between the query samples and the augmented samples: 
\begin{align} \label{eq:recon_loss_pos}
\begin{array}{l}
    \mathcal{L}_{aug}^{p} = 
    \mathbb{E}_{\scriptscriptstyle \{G, x_{\scriptscriptstyle q}\}\sim X_{\scriptscriptstyle q}} \left[\:\left\Vert G\left( e_{\scriptscriptstyle I, p}\:e_{\scriptscriptstyle A, q} \right) 
    - x_{\scriptscriptstyle q^*}\right\Vert_{1}\: \right] \vspace{0.3cm}\\
    \hspace{1cm}\:+\:  \mathbb{E}_{\scriptscriptstyle \{G, x_{\scriptscriptstyle p}\}\sim X_{\scriptscriptstyle q}} \left[\:\left\Vert G\left(
    e_{\scriptscriptstyle I, q}\:e_{\scriptscriptstyle A, p}
    \right) 
    - x_{\scriptscriptstyle p^*}\right\Vert_{1}\: \right] \vspace{0.3cm}\\
    \hspace{1cm}\:+\:
    \mathbb{E}_{\scriptscriptstyle \{G, x_{\scriptscriptstyle q}\}\sim X_{\scriptscriptstyle q}} \left[\:\left\Vert G \left( e_{\scriptscriptstyle I, q}\:e_{\scriptscriptstyle A, q} \right) 
    - x_{\scriptscriptstyle q^*}\right\Vert_{1}\: \right],
\end{array}\end{align}
where $X_q$ represents the distribution of samples belonging to the query person, and $x_{\scriptscriptstyle q^*}$ is the grayscale image of $x_{\scriptscriptstyle q}$. In practice, we use the grayscale image as an alternative to $x_{\scriptscriptstyle q}$ to reduce the impact of color on the similarity measure. 

\begin{figure*}[!t]
    \centering
    \renewcommand{\tabcolsep}{0.77mm}
    \begin{tabular}{cccccc|cccccc}
    \includegraphics[width=0.05\textwidth]{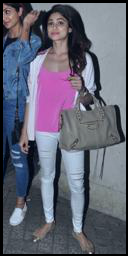} &  
    \includegraphics[width=0.05\textwidth]{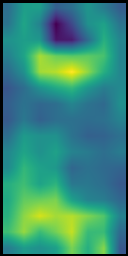} & 
    \includegraphics[width=0.05\textwidth]{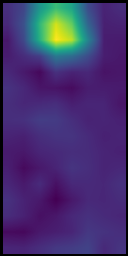} & 
    \includegraphics[width=0.05\textwidth]{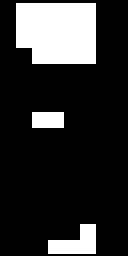} & 
    \includegraphics[width=0.05\textwidth]{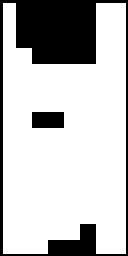} & 
    \includegraphics[width=0.05\textwidth]{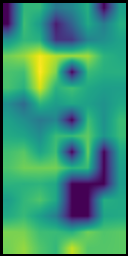} \hspace{0.05cm} & \hspace{0.05cm}
    \includegraphics[width=0.05\textwidth]{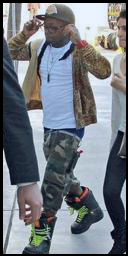}  &  
    \includegraphics[width=0.05\textwidth]{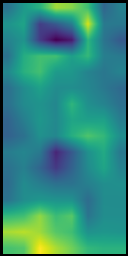} & 
    \includegraphics[width=0.05\textwidth]{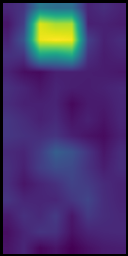} & 
    \includegraphics[width=0.05\textwidth]{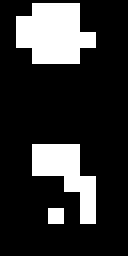} & 
    \includegraphics[width=0.05\textwidth]{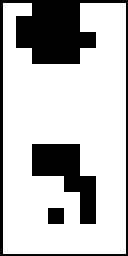} & 
    \includegraphics[width=0.05\textwidth]{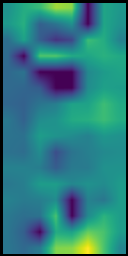} \\
    $x_{\scriptscriptstyle q}$ & {\small $E_b(x_{\scriptscriptstyle q})$} & {\small $E_g(x_{\scriptscriptstyle q})$} & $m_{\scriptscriptstyle I}^{\scriptscriptstyle q}$ & $m_{\scriptscriptstyle A}^{\scriptscriptstyle q}$ & $\tilde{x}_{\scriptscriptstyle n}^{\scriptscriptstyle q}$ & 
    $x_{\scriptscriptstyle q}$ & {\small $E_b(x_{\scriptscriptstyle q})$} & {\small $E_g(x_{\scriptscriptstyle q})$} & $m_{\scriptscriptstyle I}^{\scriptscriptstyle q}$ & $m_{\scriptscriptstyle A}^{\scriptscriptstyle q}$ & $\tilde{x}_{\scriptscriptstyle n}^{\scriptscriptstyle q}$ \vspace{0.1cm} \\
    \includegraphics[width=0.05\textwidth]{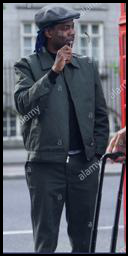} &  
    \includegraphics[width=0.05\textwidth]{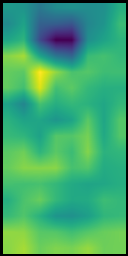} & 
    \includegraphics[width=0.05\textwidth]{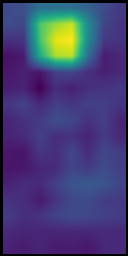} & \includegraphics[width=0.05\textwidth]{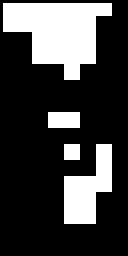}
    & \includegraphics[width=0.05\textwidth]{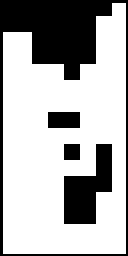} & \includegraphics[width=0.05\textwidth]{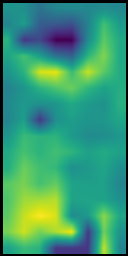} \hspace{0.05cm} & \hspace{0.05cm} 
    \includegraphics[width=0.05\textwidth]{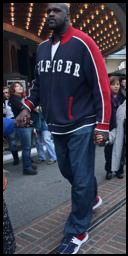} &  
    \includegraphics[width=0.05\textwidth]{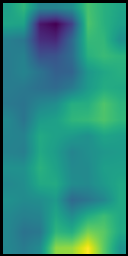} & 
    \includegraphics[width=0.05\textwidth]{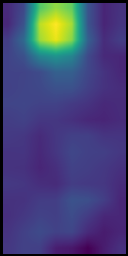} & \includegraphics[width=0.05\textwidth]{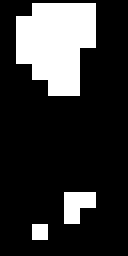}
    & \includegraphics[width=0.05\textwidth]{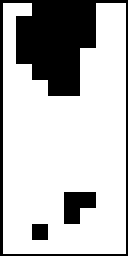} & \includegraphics[width=0.05\textwidth]{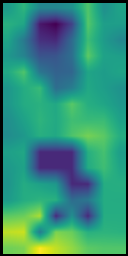} \\
    $x_{\scriptscriptstyle n}$ & {\small $E_b(x_{\scriptscriptstyle n})$} & {\small $E_g(x_{\scriptscriptstyle n})$} & $m_{\scriptscriptstyle I}^{\scriptscriptstyle n}$ & $m_{\scriptscriptstyle A}^{\scriptscriptstyle n}$ & $\tilde{x}_{\scriptscriptstyle q}^{\scriptscriptstyle n}$ &
    $x_{\scriptscriptstyle n}$ & {\small $E_b(x_{\scriptscriptstyle n})$} & {\small $E_g(x_{\scriptscriptstyle n})$} & $m_{\scriptscriptstyle I}^{\scriptscriptstyle n}$ & $m_{\scriptscriptstyle A}^{\scriptscriptstyle n}$ & $\tilde{x}_{\scriptscriptstyle q}^{\scriptscriptstyle n}$ \vspace{0.1cm} \\
    \end{tabular}
    \caption{The examples of pseudo ground truths. The pseudo ground truths are generated by mingling features with an id-relevant indicator and features with an id-irrelevant indicator. The gradient does not flow to the connections where both $m_I^q$ and $m_A^n$ or both $m_I^n$ and $m_A^q$ are zero. \vspace{-0.2cm}} \label{fig:pseudo-ground-truth}
\end{figure*}

\subsubsection{Negative Feature Augmentation}
Given a subset $\left( x_{\scriptscriptstyle q}, \: x_{\scriptscriptstyle n}\right)$ of triplet, we want the model to learn a feature embedding space, where people of different identities but similar appearances and a person of various appearances are well separable. To create such hard samples, two kinds of augmentation with negative samples are considered by swapping id-relevant embeddings with id-irrelevant embeddings. 
As a result, the generative module generates two augmentations $G\left( e_{\scriptscriptstyle I, q},\: e_{\scriptscriptstyle A, n} \right)$ and $G\left( e_{\scriptscriptstyle I, n},\: e_{\scriptscriptstyle A, q} \right)$ in the feature embedding space. However, there are no comparative references for the augmentations to measure reconstruction capability. We deal with this absence of a ground-truth problem by creating pseudo-ground-truths originating from class activation maps~\cite{zhou2016learning}. To obtain class activation maps, we instantiate a fully-connected layer followed by global average pooling to the output of the backbone network $E_{b}$. The fully-connected layer is trained using the conventional cross-entropy function $\mathcal{L}_{cam}$:
\begin{align}\label{eq:cam_loss}
    \mathcal{L}_{cam} = -\frac{1}{N_b}\sum\limits_{j}^{N_b} y_{\scriptscriptstyle q, j} \log\left( p\left( y_{\scriptscriptstyle q, j} | E(x_{\scriptscriptstyle q, j}) \right)\right), 
\end{align}
Note that we distinguish the loss $\mathcal{L}_{cam}$ denoting classification loss on pseudo-label generation and reconstruction from the loss $\mathcal{L}_{cls}$ in \eref{eq:cls_loss_feature} denoting the classification loss on the feature extraction module~(See \fref{fig:overall_architecture}). 

A class activation map for a particular category indicates the discriminative regions used for identifying that category~\cite{zhou2016learning}. Based on this attention mechanism, we can identify the most represented id-relevant features by projecting back the weights of the fully-connected layer onto the feature embedding and localizing the regions having high-intensity values. Opposite to id-relevant features, id-irrelevant features are presumably selected as the features that less contribute to classifying identities. Two indicators are then defined as $m_{\scriptscriptstyle I}^{*} = u\left( E_{g}( x_{\scriptscriptstyle *} )- \tau \right)$ for identifying id-relevant features and $m_{\scriptscriptstyle A}^{*} = u\left( \tau - E_{g}( x_{\scriptscriptstyle *} ) \right)$ for id-irrelevant features, where $E_{g}(\cdot)$ produces the class activation map, $\tau$ is the average value of the elements of the class activation map, and $u(\cdot)$ is the unit step function. Using the indicators, we create pseudo-ground-truths by re-entangling the decomposed id-relevant features of queries and id-irrelevant features of negatives, and vice versa:
\begin{align} \label{eq:pseudo_ground_truth}
\begin{array}{l}
    \widetilde{x}_{\scriptscriptstyle n}^{\scriptscriptstyle q} = m_{\scriptscriptstyle I}^{q} \otimes E_{b}( x_{\scriptscriptstyle q}) + \left(m_{\scriptscriptstyle A}^{q} \cap m_{\scriptscriptstyle A}^{n}\right) \otimes E_{b}( x_{\scriptscriptstyle n}) \\
    \widetilde{x}_{\scriptscriptstyle q}^{\scriptscriptstyle n} = m_{\scriptscriptstyle I}^{n} \otimes E_{b}( x_{\scriptscriptstyle n}) + \left(m_{\scriptscriptstyle A}^{n} \cap m_{\scriptscriptstyle A}^{q}\right) \otimes E_{b}( x_{\scriptscriptstyle q}),
\end{array}\end{align}
where operation $\otimes$ is the element-wise product, and operation $\cap$ denotes the intersection of two indicators. The feature re-entanglements in \eref{eq:pseudo_ground_truth} are designed to create feature maps that maintain the id-relevant features of a single person without any contamination by id-irrelevant features from other identities. That is, we exclude the regions relevant to either query or negative identities~(\ie $\left(m_{\scriptscriptstyle I}^{q} \cup m_{\scriptscriptstyle I}^{n}\right)^{c} = m_{\scriptscriptstyle A}^{q} \cap m_{\scriptscriptstyle A}^{n}$) when localizing id-irrelevant features, where operations $\cup$ and $c$ denote the union of two indicators and the complement of a set. The examples of the generated ground-truth feature maps are shown in \fref{fig:pseudo-ground-truth}. 

Like \eref{eq:recon_loss_pos}, the reconstruction loss measures the similarity between the pseudo-ground-truth feature map $\widetilde{x}$ and the augmented features:  
\begin{align} \label{eq:recon_loss_neg}
\begin{array}{l}
    \mathcal{L}_{aug}^{n} = 
    \mathbb{E}_{\scriptscriptstyle \{ \widetilde{G}, \widetilde{x}_{\scriptscriptstyle q}\}\sim \widetilde{X}_{\scriptscriptstyle q}} \left[\:\left\Vert \widetilde{G}\left(e_{\scriptscriptstyle I, q},\:e_{\scriptscriptstyle A, n}
    \right)
    - \widetilde{x}_{\scriptscriptstyle n}^{\scriptscriptstyle q} \right\Vert_{1} \right] \vspace{0.3cm}\\
     \hspace{1cm} + \:\mathbb{E}_{\scriptscriptstyle \{ \widetilde{G}, \widetilde{x}_{\scriptscriptstyle n}\}\sim \widetilde{X}_{\scriptscriptstyle n}} \left[\: \left\Vert \widetilde{G}\left( e_{\scriptscriptstyle I, n},\: e_{\scriptscriptstyle A, p}
    \right)
    - \widetilde{x}_{\scriptscriptstyle q}^{\scriptscriptstyle n} \right\Vert_{1} \right],
\end{array}\end{align}
where $\widetilde{X}_q$ and $\widetilde{X}_n$ represents the feature distribution of samples belonging to the query person and the negative person, respectively. 

Finally, the total reconstruction loss $\mathcal{L}_{rec}$ is the weighted sum of \eref{eq:recon_loss_pos},  \eref{eq:recon_loss_neg}, and \eref{eq:cam_loss},  as follows:
\begin{align}\label{eq:rec_loss}
    \mathcal{L}_{rec} = \lambda_{aug}^{p}\mathcal{L}_{aug}^{p} + \lambda_{aug}^{n}\mathcal{L}_{aug}^{n} \:+\: \lambda_{cam}\mathcal{L}_{cam},
\end{align}
where $\lambda_{aug}^{p}$, $\lambda_{aug}^{n}$ and $\lambda_{cam}$ are the weighting factors. The reconstruction loss in \eref{eq:rec_loss} encourages that the proposed framework consistently forms a single cluster of the embeddings of a person from both hard positive and hard negative samples. The examples of re-entangled features are shown in \fref{fig:mix}, and the examples of activation maps are in \fref{fig:activation}. 
\vspace{0.2cm}

\section{Experiments} \label{sec:results}
In this section, we validate the proposed framework described in \sref{sec:proposed}. \sref{sec:implementation} provides the implementation details of the proposed framework, and \sref{sec:datasets} describes five long-term benchmark datasets used for evaluation. \sref{sec:comparisons} shows the performance comparisons of the proposed framework against the current state-of-art models and discusses the results. In \sref{sec:ablation_study}, various ablation studies are conducted to understand the contributions of the main components of the proposed framework.

\subsection{Implementation Details}
\label{sec:implementation}
The proposed framework is implemented using PyTorch. We adopt pre-trained DenseNet-121~\cite{Huang_2017_CVPR} deleting the first pooling layer as the backbone feature extractor $E_b$, and Efficient-CapsNet~\cite{mazzia2021efficient} as the feature separator $E_s$ with squash activation function~\cite{sabour2017dynamic} and capsule dropout~\cite{8481393}. The style-based generator~\cite{Karras_2019_CVPR} is adopted as the image generator $G$. For generating class activation maps in $E_g$, we connect a global average pooling and an additional fully connected layer after $E_b$. 

In the following descriptions, channel $\times$ height $\times$ width to denote the size of tensors, all input images used in our method are resized $3 \times 256 \times 128$, $E_b\left(x\right)$ is $1024 \times 16 \times 8$, $e_{\scriptscriptstyle I}$ is $1920 \times 1 \times 1$, and $e_{\scriptscriptstyle A}$ is $128 \times 1 \times 1$. The Adam optimizer~\cite{kingma2014adam} is used to optimize our method with a learning rate of $0.0002$ and $\left(\beta_1, \beta_2\right) = (0.9, 0.999)$. The weighting factors $\mathcal{L}_{id}$ and $\mathcal{L}_{rec}$ in \eref{eq:total_loss_2} are set to $1$. The factors $\mathcal{L}_{cls}$, $\mathcal{L}_{tri}$, and $\mathcal{L}_{sim}$ in \eref{eq:id_loss} are set to $0.05$, $1$, and $0.5$, and the weighting factors $\mathcal{L}_{aug}^{p}$, $\mathcal{L}_{aug}^{n}$ and $\mathcal{L}_{cam}$ are set to $0.0001$, $0.0001$ and $1$, respectively. The margin $m$ in \eref{eq:L_tri} sets to $0.9$. 

In training, input images are randomly augmented by grayscale images with a probability of $0.1$. In testing, we use horizontally flipped images with the untransformed images, and $\alpha$ is set to $0.55$ for similarity distance calculation.
 
\begin{table*}
    \centering
    \caption{Comparisons of performance on Celeb-reID, Celeb-reID-light, PRCC, LTCC, and VC-Clothes datasets in terms of Rank1/Rank5 classification accuracy and mAP.} 
    \label{tab:benchmark_results}
    \renewcommand{\tabcolsep}{0.77mm}
    {\renewcommand{\arraystretch}{1.1}
    \begin{tabular}{l|ccc|ccc|cc|cc|cc}
        \hline
        \multirow{3}{*}{Method} & \multicolumn{3}{c|}{\multirow{2}{*}{Celeb-reID~\cite{Huang_2020_TCSVT}}} &
        \multicolumn{3}{c|}{\multirow{2}{*}{Celeb-reID-light~\cite{huang2019celebrities}}} & 
        \multicolumn{4}{c|}{LTCC~\cite{qian2020long}} & 
        \multicolumn{2}{c}{\multirow{2}{*}{VC-Clothes~\cite{wan2020person}}}\\ \cline{8-11}
        & & & & & & & \multicolumn{2}{c|}{Standard} & \multicolumn{2}{c|}{Clothe Changing}\\ \cline{2-13}
        & Rank1 & Rank5 & mAP & Rank1 & Rank5 & mAP & Rank1 & mAP & Rank1 & mAP & Rank1 & mAP \\ \hline \hline
        
        MDLA~\cite{Qian_2017_ICCV} & - & - & - & - & - & - & - & - & - & - & 88.9 & 76.8 \\
        PCB~\cite{Sun_2018_ECCV} & 37.1 & 57.0 & 8.2 & - & - & - & 65.1 & 30.6 & 23.5 & 10.0 & 87.7 & 74.6 \\
        3APF~\cite{wan2020person} & - & - & - & - & - & - & - & - & - & - & 90.2 & 82.1 \\
        Part-aligned~\cite{Suh_2018_ECCV} & - & - & - & - & - & - & - & - & - & - & 90.5 & 79.7 \\
        MLFN~\cite{Chang_2018_CVPR} & 41.4 & 54.7 & 6.0 & 10.6 & 31.0 & 6.3  & - & - & - & -  & - & - \\
        IDE~\cite{Zheng_2017_CVPR} & 42.9 & 56.4 & 5.9 & 10.5 & 24.8 & 5.3 & - & - & - & -  & - & - \\
        ResNet-Mid~\cite{yu2017devil} & 43.3 & 54.6 & 5.8 & 10.3 & 28.0 & 6.0 & - & - & - & -  & - & - \\
        HACNN~\cite{Li_2018_CVPR} & 47.6 & 63.3 & 9.5 & 16.2 & 42.8 & 11.5 & 60.2 & 26.7 & 21.6 & 9.3 & - & - \\
        MGN~\cite{wang2018learning} & 49.0 & 64.9 & 10.8 & 21.5 & 47.4 & 13.9 & - & - & - & -  & - & - \\
        CESD~\cite{qian2020long} & 50.9 & 66.3 & 9.8 & - & - & - & 71.4 & 34.3 & 26.2 & 12.4 & - & - \\
        GI-ReID~\cite{jin2021cloth} & - & - & - & - & - & - & 73.6 & 36.1 & 28.1 & 13.2 & - & - \\
        ReIDCaps~\cite{Huang_2020_TCSVT} & 51.2 & 65.4 & 9.8 & 20.3 & 48.2 & 11.2 & {60.5} & {26.8} & 22.2 & 10.5  & 79.2 & 59.4 \\
        AFD-Net~\cite{xuadversarial} & 52.1 & 66.1 & 10.6 & 22.2 & 51.0 & 11.3 & - & - & - & -  & - & - \\
        FSAM~\cite{hong2021fine} & - & - & - & - & - & - & 73.2 & 35.4 & 38.5 & 16.2  & - & - \\
        LaST~\cite{shu2021large} & 54.4 & - & 11.8 & 29.0 & - & 16.3 & 71.8 & 34.1 & 34.4 & 14.7  & \textbf{92.3} & 84.9 \\
        RCSANet~\cite{huang2021clothing} & 55.6 & - & 11.9 & 29.5 & - & 16.7 & - & - & - & - & - & - \\ \hline
        \textbf{Ours} & \textbf{56.0} & \textbf{70.3} & \textbf{14.2} & \textbf{36.0} & \textbf{65.3} & \textbf{20.0} & \textbf{76.7} & \textbf{47.2} & \textbf{40.3} & \textbf{24.0} & \textbf{92.3} & \textbf{87.4} \\
        
        \hline
    \end{tabular}  
    } \vspace{-0.3cm}
\end{table*}

\subsection{Datasets} \label{sec:datasets}
We evaluate our framework on four large-scale benchmark datasets under clothe changing: Celeb-reID~\cite{Huang_2020_TCSVT}, Celeb-reID-light~\cite{huang2019celebrities}, long-term cloth-changing~(LTCC) dataset~\cite{qian2020long}, and virtually changing-clothes~(VC-Clothes)~\cite{wan2020person}.

\noindent \textbf{Celeb-reID}~\cite{Huang_2020_TCSVT} consists of 34,186 images of 1052 identities that are crawled street snap-shots from websites. The data is split into two parts: 632 identities with 20,208 images for training and 420 identities with 13,978 images for testing. Among the test set, 2,972 images are used for query, and 11,006 images are used for the gallery. \\  
\noindent \textbf{Celeb-reID-light}~\cite{huang2019celebrities} is the light version of Celeb-reID dataset, consisting of 590 identities with 10,842 images. It also split into 490 identities for training and 100 identities for testing. For testing, 887 images are used as queries, and 934 images are used as galleries. In the Celeb-reID-light dataset, all people wear entirely different clothes, and only more than 70\% of people wear different clothes in the Celeb-reID dataset. \\ 
\noindent \textbf{LTCC}~\cite{qian2020long} is a large-scale indoor clothe changing dataset captured by 12 cameras with various environmental settings.  The training set consists of 77 identities where 46 people have cloth changes and 31 people wear the same clothes. Similarly, the testing set contains 45 people wearing different clothes and 30 people wearing the same clothes. For the standard evaluation setting, both cloth-consistent and clothe changing samples are in the test set, and there are only clothe changing samples for the clothe changing evaluation. \\ 
\noindent \textbf{VC-Clothes}~\cite{wan2020person} is a synthetic dataset using 3D human models of the GTA5 engine. It contains 512 identities of 19,060 images in 4 different scenes with significant clothes changes: 256 identities for training and the other 256 for testing. For testing, 1,020 images are used as queries and 8,591 others are used as the gallery, and 9,449 images are used for training. Each person wears the same clothes in Cameras 2 and 3, and each person wears different clothes in Cameras 3 and 4.
\subsection{Comparison with State-of-the-Art}\label{sec:comparisons}
We compare the proposed model against the current-state-of-art models, including MDLA~\cite{Qian_2017_ICCV}, PCB~\cite{Sun_2018_ECCV}, Part-aligned~\cite{Suh_2018_ECCV}, MLFN~\cite{Chang_2018_CVPR}, IDE~\cite{Zheng_2017_CVPR}, ResNet-Mid~\cite{yu2017devil}, HACNN~\cite{Li_2018_CVPR}, MGN~\cite{wang2018learning}, CESD~\cite{qian2020long}, GI-ReID~\cite{jin2021cloth}, ReIDCaps~\cite{Huang_2020_TCSVT}, AFD-Net~\cite{xuadversarial}, FSAM~\cite{hong2021fine}, LaST~\cite{shu2021large}, and RCSANet~\cite{huang2021clothing}.
The comparative results of the proposed framework on the benchmark datasets are summarized in \tref{tab:benchmark_results} in terms of mean average precision~(mAP) and rank-$k$ accuracy~\cite{zhong2017re}. Overall, the proposed model shows a better performance of all compared models, demonstrating the effectiveness of appearance changes.

As demonstrated in the comparisons, the proposed framework outperforms all the compared methods on Celeb-reID and Celeb-reID-light datasets. In particular, the Rank1 improves by about relatively $22.0\%$ on Celeb-reID-light and $0.7\%$ on Celeb-reID compared to the current-state-of-art model. The mAP improves by $19.8\%$ on Celeb-reID-light and $19.3\%$ on Celeb-reID. This significant gain on the datasets verifies that the proposed framework robustly distinguishes identities in appearance and environmental variations. 

For the LTCC dataset, we conduct experiments on both the standard-setting and the clothe changing setting since they all include clothe changing samples in testing. The proposed model also outperforms the compared models on this dataset for both standard-setting and clothe changing settings. The Rank1 and mAP improve by about $4.21\%$ and $30.7\%$ on the standard-setting. The performance on the clothe changing setting improves by $4.68\%$ in Rank1 and $48.1\%$ in mAP. As shown in the results of the two evaluations, significant performance degradation is observed in the clothe changing test, which implies intra-identity variations measure is essential to improve the model's generalization performance as we propose. Furthermore, the higher performance gain on the clothe changing setting indicates that the proposed framework is beneficial to identify a person in appearance variations. We also compare the performance of the proposed model on the clothe changing samples of the synthesized dataset, VC-Clothes -- an improvement of about $1.99\%$ in Rank1 and $2.94\%$ in mAP. 
\begin{figure*}[!t]
    \centering
    \renewcommand{\tabcolsep}{1mm}
    \begin{tabular}{@{}c@{~}|c@{~}|c@{~}|c@{~}|c@{~}|c@{}}
        & \footnotesize Query & \footnotesize $\mathcal{L}_{cls}+\mathcal{L}_{tri}$ & \footnotesize $\mathcal{L}_{cls}+\mathcal{L}_{tri}+\mathcal{L}_{sim}$ & \footnotesize $\mathcal{L}_{cls}+\mathcal{L}_{tri}+\mathcal{L}_{rec}$ & \footnotesize $\mathcal{L}_{cls}+\mathcal{L}_{tri}+\mathcal{L}_{sim}+\mathcal{L}_{rec}$ \\ \hline
        \rotatebox{90}{Celeb-reID} &
        \includegraphics[width=0.0386\textwidth]{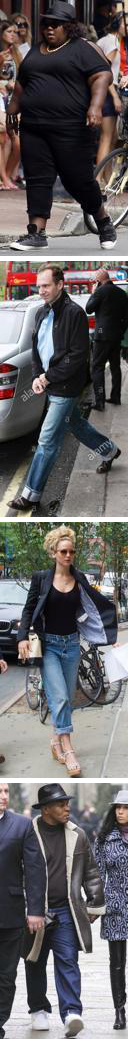} &
        \includegraphics[width=0.16\textwidth]{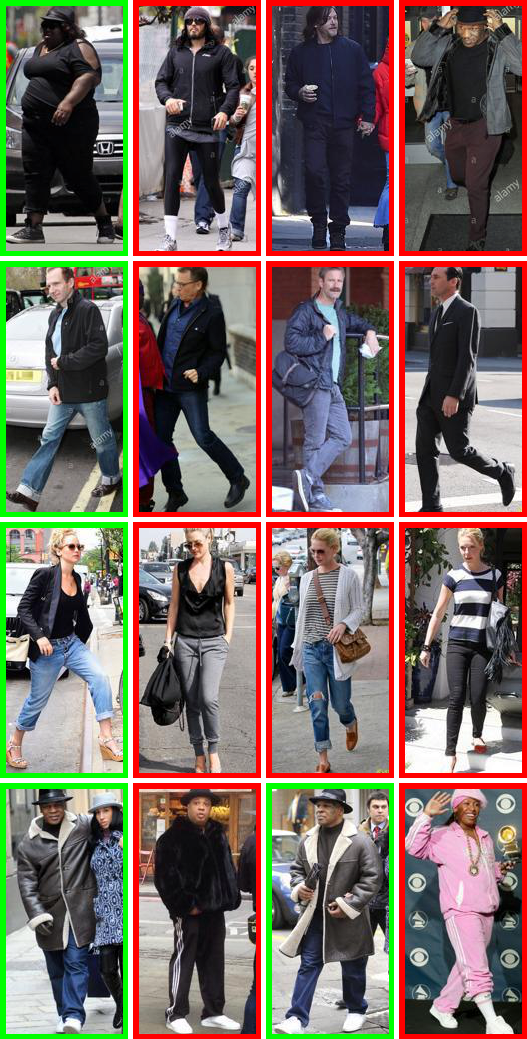} &
        \includegraphics[width=0.16\textwidth]{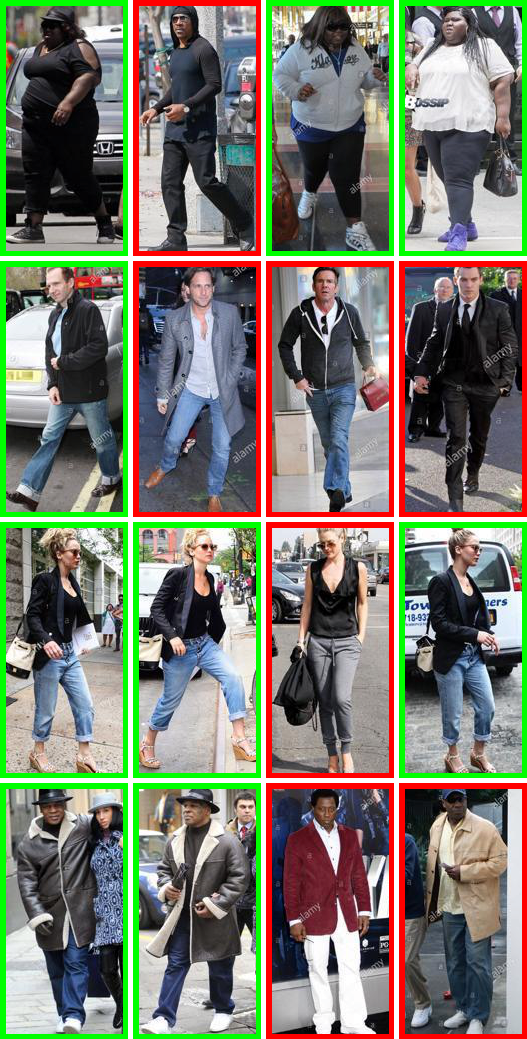} &
        \includegraphics[width=0.16\textwidth]{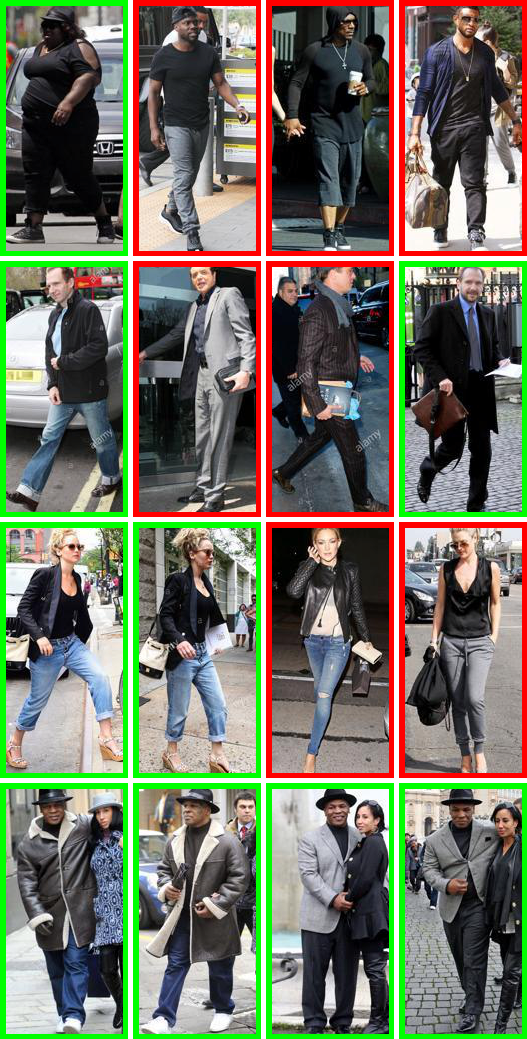} &
        \includegraphics[width=0.16\textwidth]{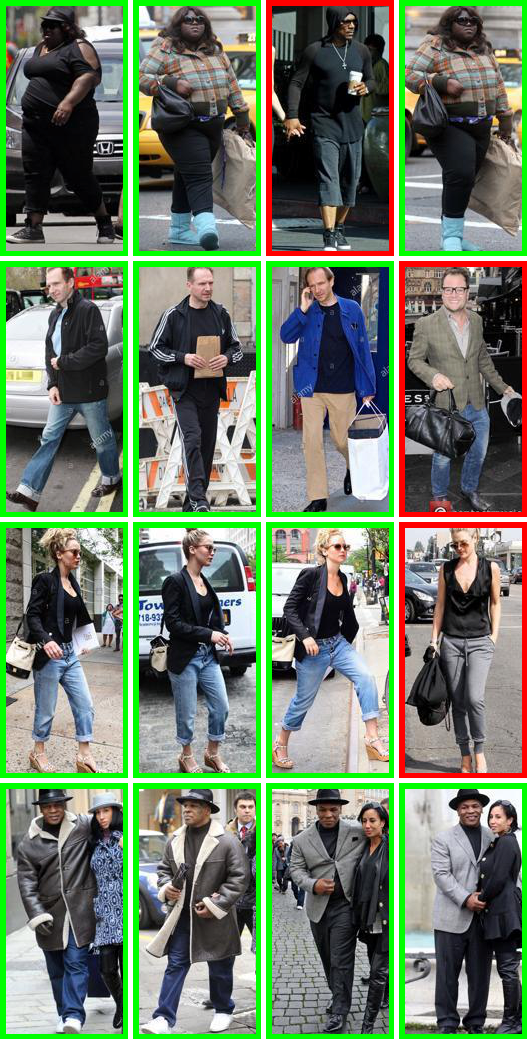}
        \\ \hline
        \rotatebox{90}{LTCC~(Clothes-Changing)} &
        \includegraphics[width=0.0386\textwidth]{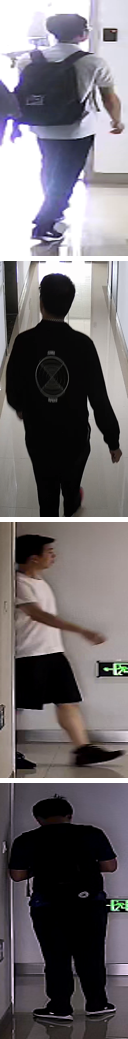} &
        \includegraphics[width=0.16\textwidth]{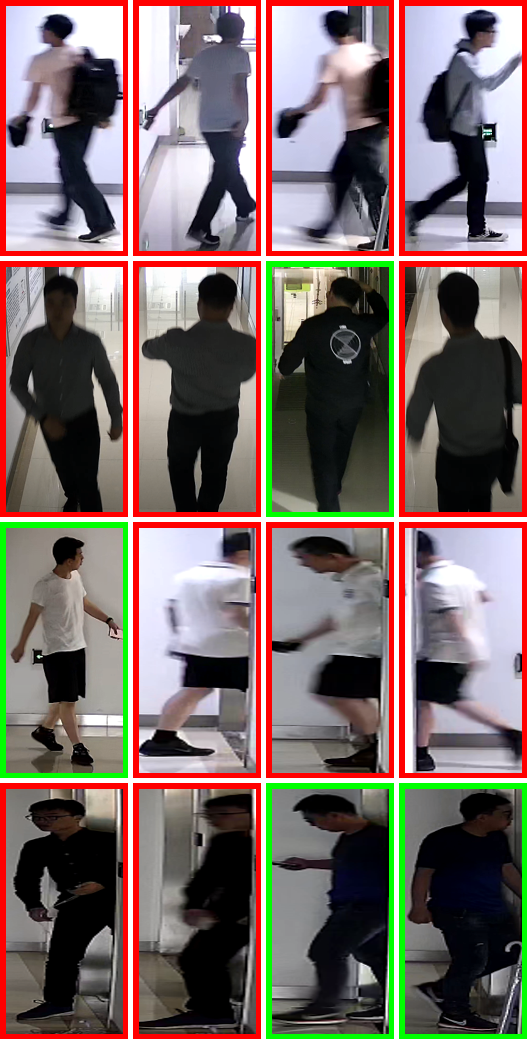} &
        \includegraphics[width=0.16\textwidth]{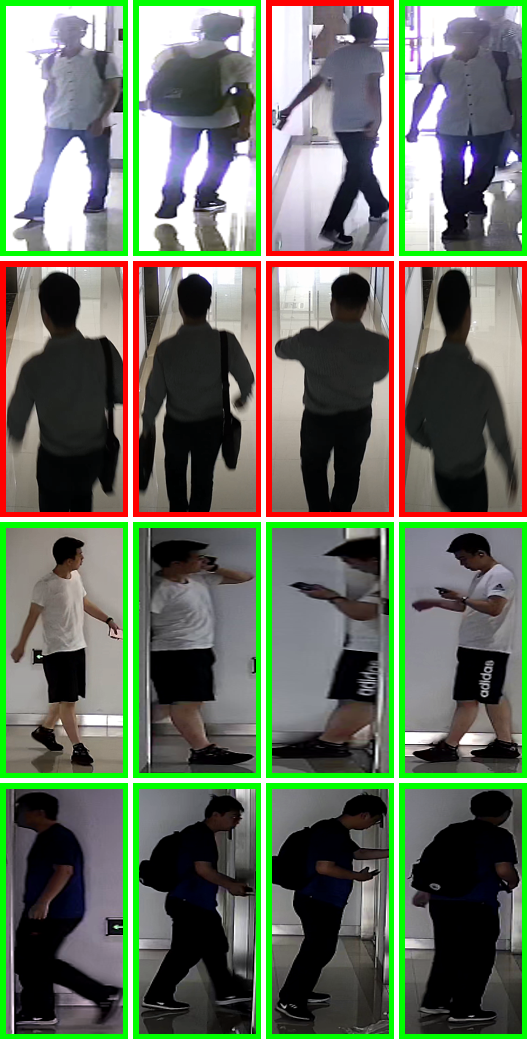} &
        \includegraphics[width=0.16\textwidth]{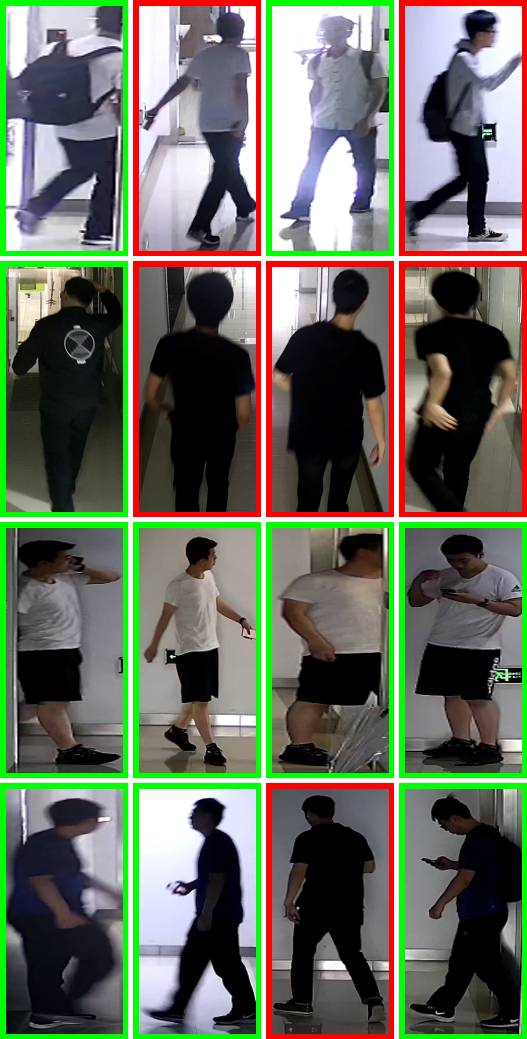} &
        \includegraphics[width=0.16\textwidth]{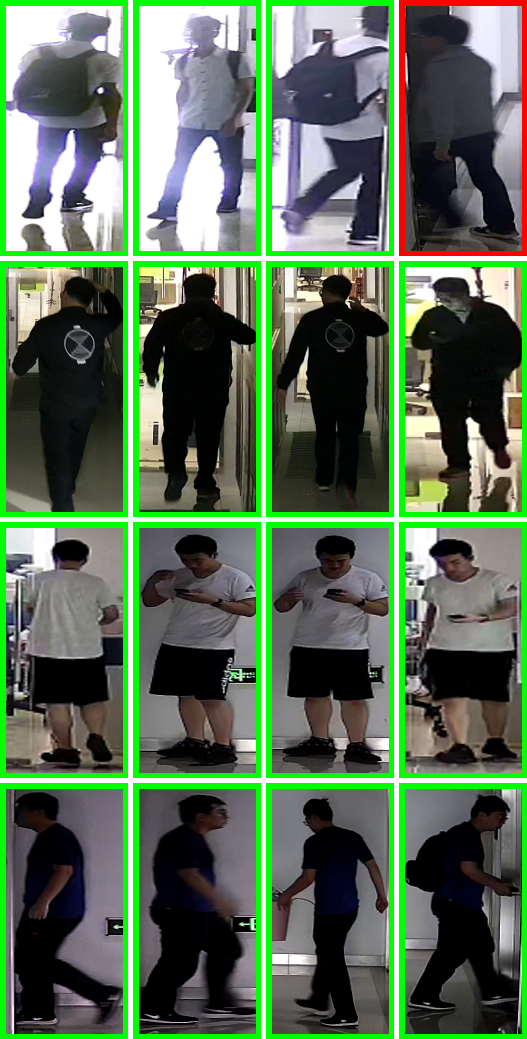}
        \\
    \end{tabular} \vspace{0.1cm}
    \caption{Examples of top-4 retrieval results with different combinations of loss functions on Celeb-reID~\cite{Huang_2020_TCSVT} and LTCC~\cite{qian2020long} datasets. Green boxes are images belonging to the same identity as the query person. Red boxes are images belonging to other people. As shown in the above examples, the proposed feature representation and feature augmentation methods help to improve person re-identification performance.} \vspace{-0.2cm}
    \label{fig:ranking}
\end{figure*}

\begin{table}
    \centering
    \caption{Ablation study of our loss function on Celeb-reID~\cite{Huang_2020_TCSVT} and Celeb-reID-light~\cite{huang2019celebrities} datasets. ($\mathcal{L}_{cam}$ is omitted in the fourth and fifth rows.)
    }
    \label{tab:ablation_loss}
    \renewcommand{\tabcolsep}{0.8mm}
    {\renewcommand{\arraystretch}{1.2}
    \begin{tabular}{@{}l|ccc|ccc}
        \hline
        \multirow{2}{*}{Method} & 
        \multicolumn{3}{c|}{Celeb-reID~\cite{Huang_2020_TCSVT}} &
        \multicolumn{3}{c}{Celeb-reID-light~\cite{huang2019celebrities}} \\ \cline{2-7}
        
         & Rank1 & Rank5 & mAP & Rank1 & Rank5 & mAP\\ \hline \hline
        
       $\mathcal{L}_{cls} + \mathcal{L}_{tri}$ & 47.5 & 63.1 & 11.2 & 30.8 & 62.3 & 18.2 \\
       $\mathcal{L}_{cls} + \mathcal{L}_{tri} + \mathcal{L}_{sim}$ & 51.3 & 65.7 & 11.7 & 30.6 & 62.9 & 18.3 \\
       $\mathcal{L}_{cls} + \mathcal{L}_{tri} + \mathcal{L}_{rec}$ & 53.7 & 68.9 & 13.6 & 32.7 & 63.7 & 18.7 \\
       $\mathcal{L}_{cls} + \mathcal{L}_{tri} + \mathcal{L}_{sim} + \mathcal{L}_{aug}^{n}$ & 54.7 & 68.9 & 14.0 & 35.2 & 64.9 & 19.6 \\
       $\mathcal{L}_{cls} + \mathcal{L}_{tri} + \mathcal{L}_{sim} + \mathcal{L}_{aug}^{p}$ & 55.0 & 69.4 & 13.7 & 33.5 & \textbf{65.3} & 19.1 \\
       $\mathcal{L}_{cls} + \mathcal{L}_{tri} + \mathcal{L}_{sim} + \mathcal{L}_{rec}$ & \textbf{56.0} & \textbf{70.3} & \textbf{14.2} & \textbf{36.0} & \textbf{65.3} & \textbf{20.0} \\ \hline
    \end{tabular}}
\end{table}

\subsection{Ablation Study} \label{sec:ablation_study}
\subsubsection{Loss Functions}
To further analyze the effectiveness of the introduced loss functions, we train our model by disabling one or more loss functions. For the analysis, we consider the combination of cross-entropy loss and triplet loss as the baseline~(``$\mathcal{L}_{cls}+\mathcal{L}_{tri}$''), comparing to five different combinations: the loss function with the proposed sample independent feature representation module~(``$\mathcal{L}_{cls}+\mathcal{L}_{tri}+\mathcal{L}_{sim}$''), the loss function with the proposed feature augmentation module and the pseudo label generation module~(``$\mathcal{L}_{cls}+\mathcal{L}_{tri}+\mathcal{L}_{rec}$''), the loss function on the proposed negative feature augmentation with pseudo label generation~(``$\mathcal{L}_{cls}+\mathcal{L}_{tri}+\mathcal{L}_{aug}^n+\mathcal{L}_{cam}$''), the loss function including the positive feature augmentation module~(``$\mathcal{L}_{cls}+\mathcal{L}_{tri}+\mathcal{L}_{aug}^p+\mathcal{L}_{cam}$''), and the loss function including both the proposed sample independent feature representation module and the feature augmentation module~(``$\mathcal{L}_{cls}+\mathcal{L}_{tri}+\mathcal{L}_{sim}+\mathcal{L}_{rec}$''). 

\begin{figure*}[!t]
    \centering
    \renewcommand{\tabcolsep}{1mm}
    \begin{tabular}{@{}ccc|ccc c@{}}
        \multicolumn{3}{c|}{Celeb-reID}& \multicolumn{3}{c}{LTCC}\\
        \includegraphics[width=0.15\textwidth]{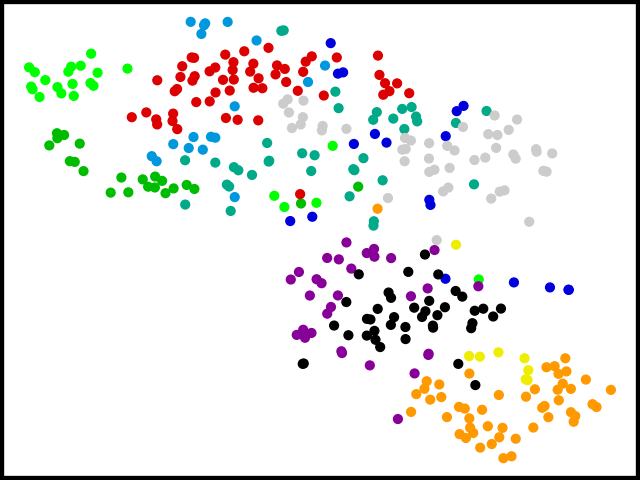} & \includegraphics[width=0.15\textwidth]{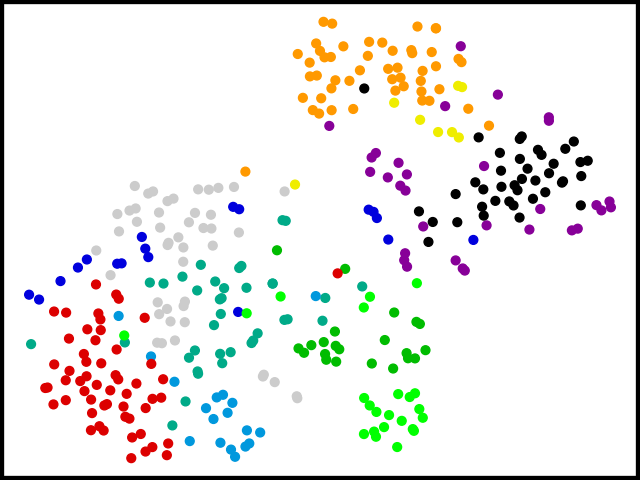} & 
        \includegraphics[width=0.15\textwidth]{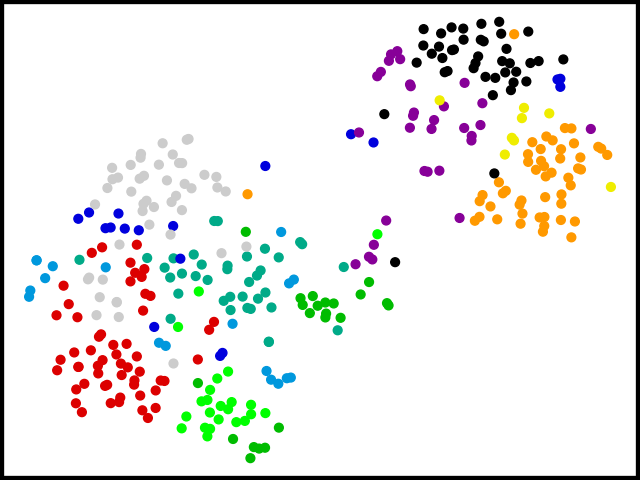} & 
        \includegraphics[width=0.15\textwidth]{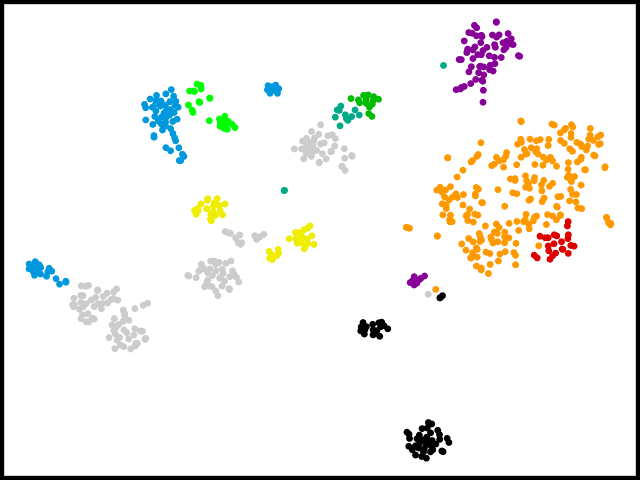} & 
        \includegraphics[width=0.15\textwidth]{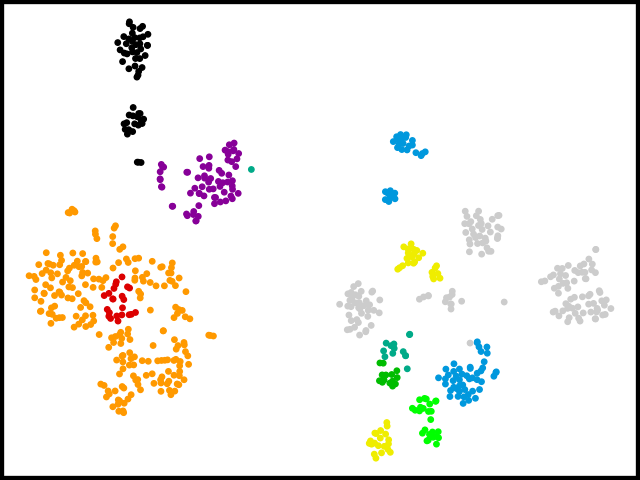} &
        \includegraphics[width=0.15\textwidth]{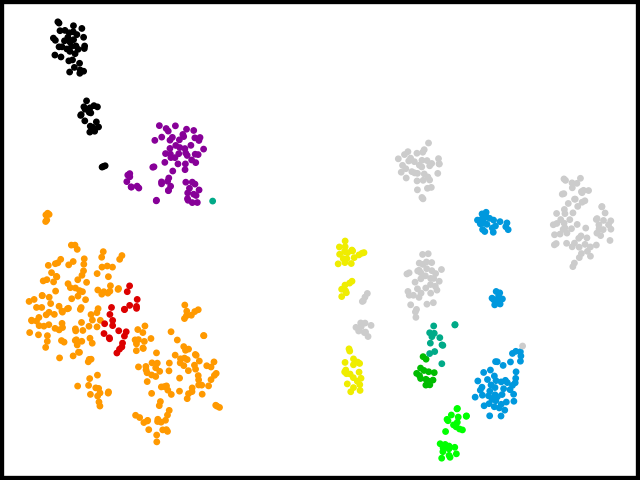} & 
        \multirow{4}{*}{\includegraphics[scale=0.4]{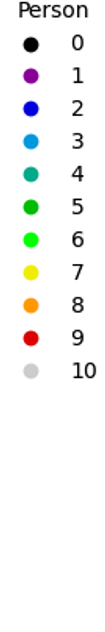}}\\
        \includegraphics[width=0.15\textwidth]{figs/tsne_cluster/celeb_05_rep.png} & 
        \includegraphics[width=0.15\textwidth]{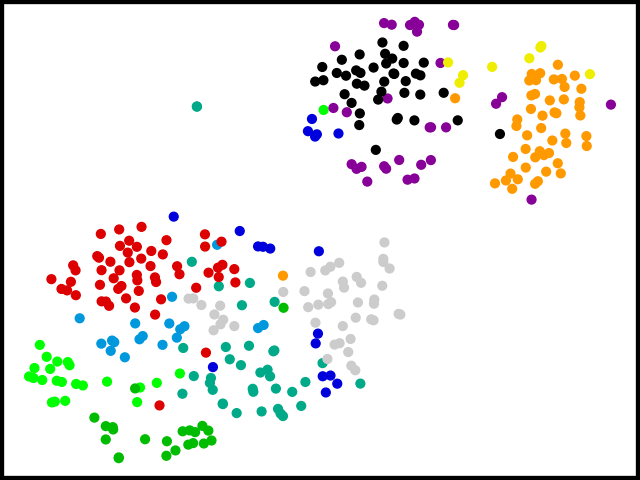} &
        \includegraphics[width=0.15\textwidth]{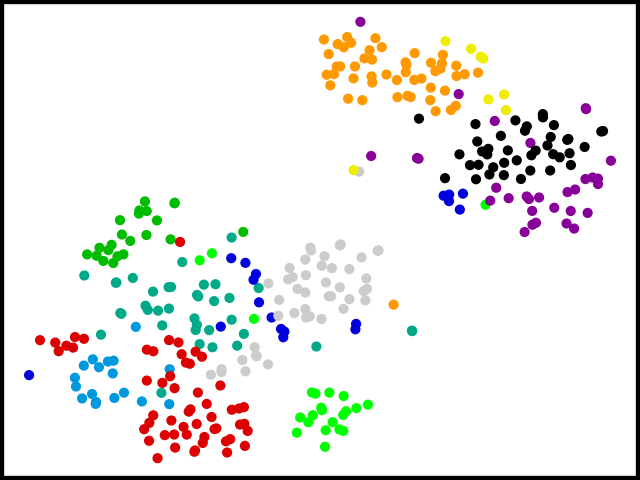} & 
        \includegraphics[width=0.15\textwidth]{figs/tsne_cluster/LTCC_rep05.png} &
        \includegraphics[width=0.15\textwidth]{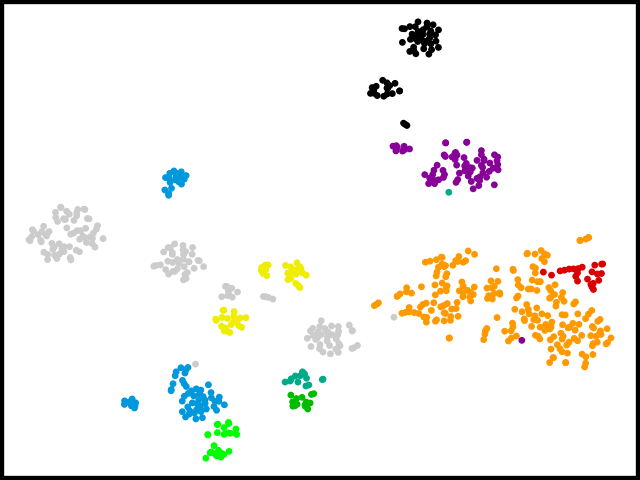} & 
        \includegraphics[width=0.15\textwidth]{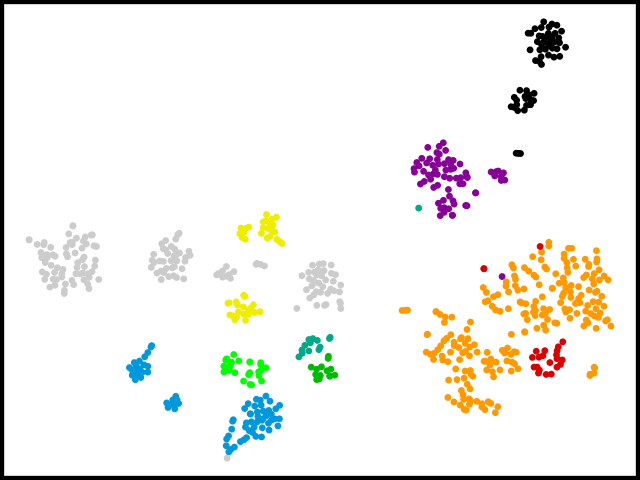} & \\
        $5^{\textrm{th}}$ epoch & $10^{\textrm{th}}$ epoch & $15^{\textrm{th}}$ epoch & $5^{\textrm{th}}$ epoch & $50^{\textrm{th}}$ epoch & $100^{\textrm{th}}$ epoch & \\
    \end{tabular}
    \caption{Visualization of the t-SNE~\cite{van2008visualizing} scatter plots on Celeb-reID~\cite{Huang_2020_TCSVT} and LTCC~\cite{qian2020long} datasets. The 11 identities are randomly selected, with 160 samples from each identity in the galleries of both datasets. The figures in the first row are the clusters of the selected identities using the model trained without sample-independent maximum discrepancy loss $\mathcal{L}_{sim}$. The figures in the second row are the clusters using the model trained with $\mathcal{L}_{sim}$. As demonstrated in the figure, the sample-independent maximum discrepancy loss with classification losses performs better clustering than without it. Note that the mAP and Rank1 accuracy~(mAP/Rank1) at the $15^{\textrm{th}}$ epoch are $(13.6\% / 53.7\%)$ for the model without $\mathcal{L}_{sim}$ and $(14.2\% /56.0\%)$ for the model with $\mathcal{L}_{sim}$ on Celeb-reID. The mAP and Rank1 accuracy~(mAP/Rank1) at the $100^{\textrm{th}}$ epoch are $(46.3\% /75.5\%)$ for the model without $\mathcal{L}_{sim}$ and $(47.2\% /76.7\%)$ for the model with $\mathcal{L}_{sim}$ on LTCC.}
    \label{fig:tsne_cluster}
\end{figure*}

\begin{figure*}[h]
    \centering
    \begin{tabular}{@{}c@{~}c@{~}c@{~}c@{~}c@{~}c@{~}|c@{~}c@{~}c@{~}c@{~}c@{~}||c@{~}c@{~}c@{~}c@{~}|c@{~}c@{~}c@{~}c@{~}}
    \parbox[t]{4mm}{\multirow{8}{*}{\rotatebox[origin=c]{90}{Identity relevant $\xleftarrow{\makebox[4cm]{}}$~~~~~~}}} \hspace{0.2cm}
    & 
    \multicolumn{18}{l}{$\xrightarrow{\makebox[4cm]{}}$ Identity irrelevant} \vspace{0.2cm}
    \\
    & & 
    \includegraphics[width=0.04\textwidth]{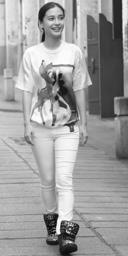} & 
    \includegraphics[width=0.04\textwidth]{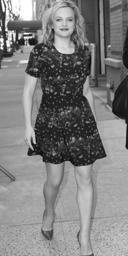} & 
    \includegraphics[width=0.04\textwidth]{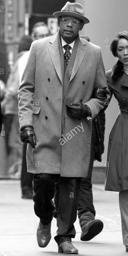} & 
    \includegraphics[width=0.04\textwidth]{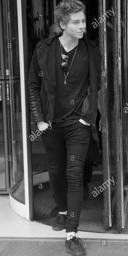} &
    &
    \includegraphics[width=0.04\textwidth]{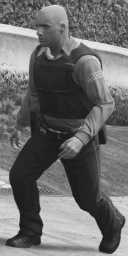} & 
    \includegraphics[width=0.04\textwidth]{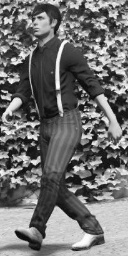} & 
    \includegraphics[width=0.04\textwidth]{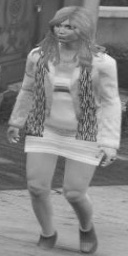} & 
    \includegraphics[width=0.04\textwidth]{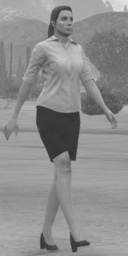} &
    & 
    \includegraphics[width=0.04\textwidth]{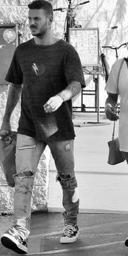} & 
    & 
    \includegraphics[width=0.04\textwidth]{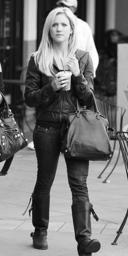} &
    & 
    \includegraphics[width=0.04\textwidth]{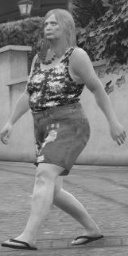} & 
    & 
    \includegraphics[width=0.04\textwidth]{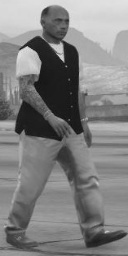}
    \\
    &
    \includegraphics[width=0.04\textwidth]{figs/mix_imgs/celeb/gray/0.jpg} & 
    \includegraphics[width=0.04\textwidth]{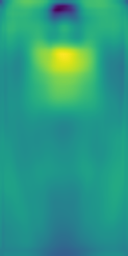} & 
    \includegraphics[width=0.04\textwidth]{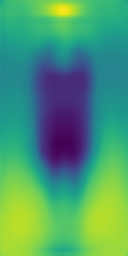} & 
    \includegraphics[width=0.04\textwidth]{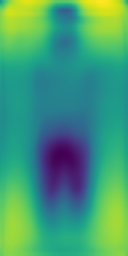} & 
    \includegraphics[width=0.04\textwidth]{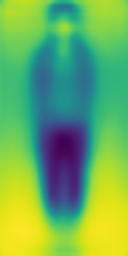} &
    \includegraphics[width=0.04\textwidth]{figs/mix_imgs/vcc/gray/0.jpg} & 
    \includegraphics[width=0.04\textwidth]{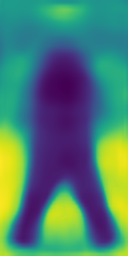} & 
    \includegraphics[width=0.04\textwidth]{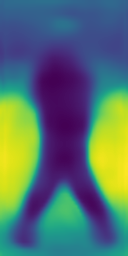} & 
    \includegraphics[width=0.04\textwidth]{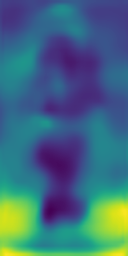} & 
    \includegraphics[width=0.04\textwidth]{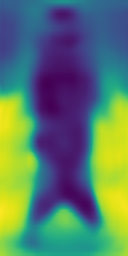} &
    \includegraphics[width=0.04\textwidth]{figs/mix_imgs/celeb_positive/gray/0.png} & 
    \includegraphics[width=0.04\textwidth]{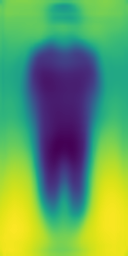} & 
    \includegraphics[width=0.04\textwidth]{figs/mix_imgs/celeb_positive/gray/4.png} & 
    \includegraphics[width=0.04\textwidth]{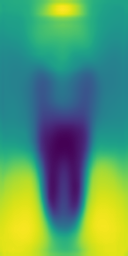} &
    \includegraphics[width=0.04\textwidth]{figs/mix_imgs/vcc_positive/gray/0.jpg} & 
    \includegraphics[width=0.04\textwidth]{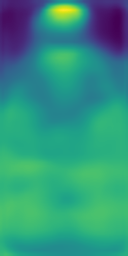} & 
    \includegraphics[width=0.04\textwidth]{figs/mix_imgs/vcc_positive/gray/4.jpg} & 
    \includegraphics[width=0.04\textwidth]{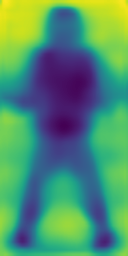}
    \\
    &
    \includegraphics[width=0.04\textwidth]{figs/mix_imgs/celeb/gray/1.jpg} & 
    \includegraphics[width=0.04\textwidth]{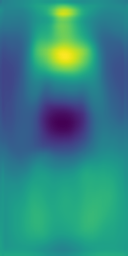} & 
    \includegraphics[width=0.04\textwidth]{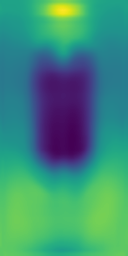} & 
    \includegraphics[width=0.04\textwidth]{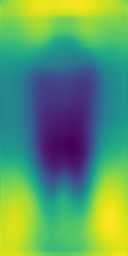} & 
    \includegraphics[width=0.04\textwidth]{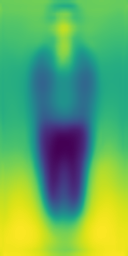} &
    \includegraphics[width=0.04\textwidth]{figs/mix_imgs/vcc/gray/1.jpg} & 
    \includegraphics[width=0.04\textwidth]{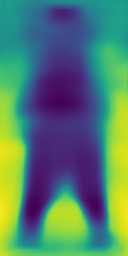} & 
    \includegraphics[width=0.04\textwidth]{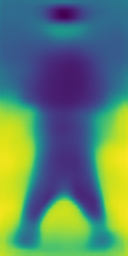} & 
    \includegraphics[width=0.04\textwidth]{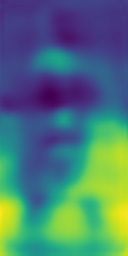} & 
    \includegraphics[width=0.04\textwidth]{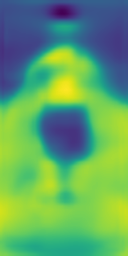} &
    \includegraphics[width=0.04\textwidth]{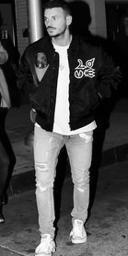} & 
    \includegraphics[width=0.04\textwidth]{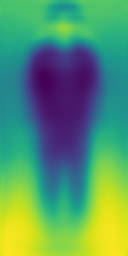} & 
    \includegraphics[width=0.04\textwidth]{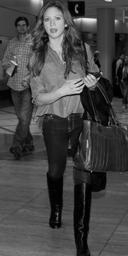} & 
    \includegraphics[width=0.04\textwidth]{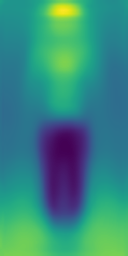} &
    \includegraphics[width=0.04\textwidth]{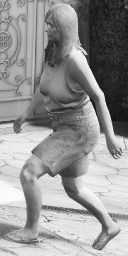} & 
    \includegraphics[width=0.04\textwidth]{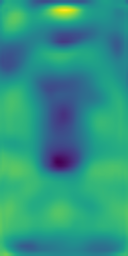} & 
    \includegraphics[width=0.04\textwidth]{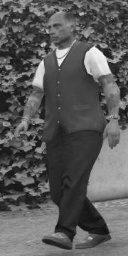} & 
    \includegraphics[width=0.04\textwidth]{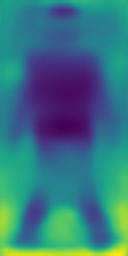}
    \\
    &
    \includegraphics[width=0.04\textwidth]{figs/mix_imgs/celeb/gray/2.jpg} & 
    \includegraphics[width=0.04\textwidth]{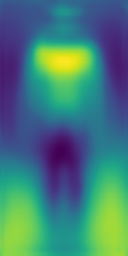} & 
    \includegraphics[width=0.04\textwidth]{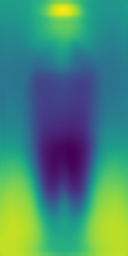} & 
    \includegraphics[width=0.04\textwidth]{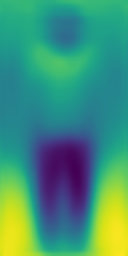} & 
    \includegraphics[width=0.04\textwidth]{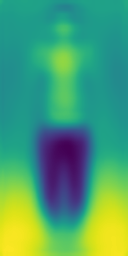} &
    \includegraphics[width=0.04\textwidth]{figs/mix_imgs/vcc/gray/2.jpg} & 
    \includegraphics[width=0.04\textwidth]{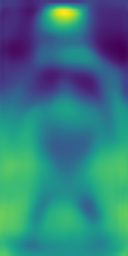} & 
    \includegraphics[width=0.04\textwidth]{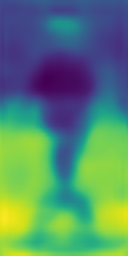} & 
    \includegraphics[width=0.04\textwidth]{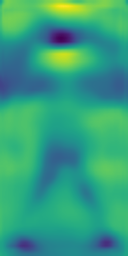} & 
    \includegraphics[width=0.04\textwidth]{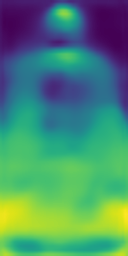} &
    \includegraphics[width=0.04\textwidth]{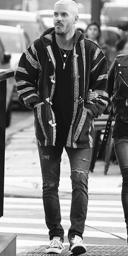} & 
    \includegraphics[width=0.04\textwidth]{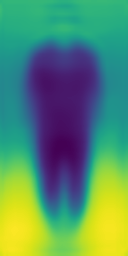} & 
    \includegraphics[width=0.04\textwidth]{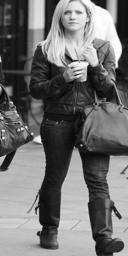} & 
    \includegraphics[width=0.04\textwidth]{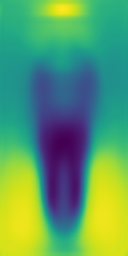} & 
    \includegraphics[width=0.04\textwidth]{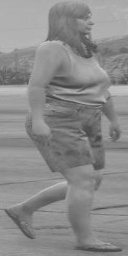} & 
    \includegraphics[width=0.04\textwidth]{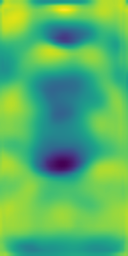} & 
    \includegraphics[width=0.04\textwidth]{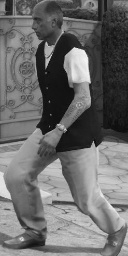} & 
    \includegraphics[width=0.04\textwidth]{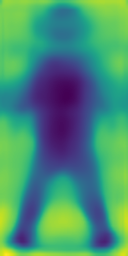}
    \\
    &
    \includegraphics[width=0.04\textwidth]{figs/mix_imgs/celeb/gray/3.jpg} & 
    \includegraphics[width=0.04\textwidth]{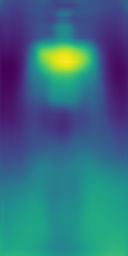} & 
    \includegraphics[width=0.04\textwidth]{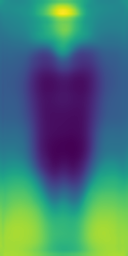} & 
    \includegraphics[width=0.04\textwidth]{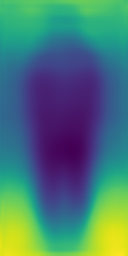} & 
    \includegraphics[width=0.04\textwidth]{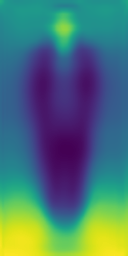} &
    \includegraphics[width=0.04\textwidth]{figs/mix_imgs/vcc/gray/3.jpg} & 
    \includegraphics[width=0.04\textwidth]{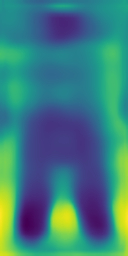} & 
    \includegraphics[width=0.04\textwidth]{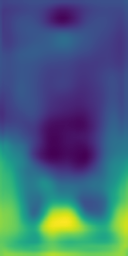} & 
    \includegraphics[width=0.04\textwidth]{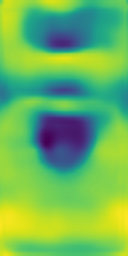} & 
    \includegraphics[width=0.04\textwidth]{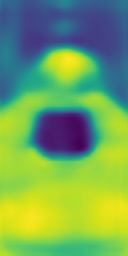} &
    \includegraphics[width=0.04\textwidth]{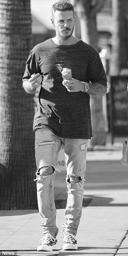} & 
    \includegraphics[width=0.04\textwidth]{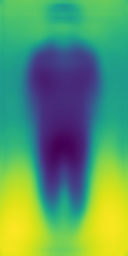} & 
    \includegraphics[width=0.04\textwidth]{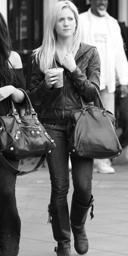} & 
    \includegraphics[width=0.04\textwidth]{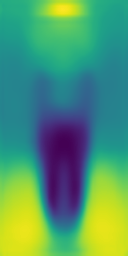} &
    \includegraphics[width=0.04\textwidth]{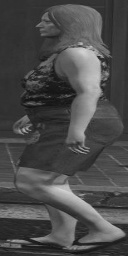} & 
    \includegraphics[width=0.04\textwidth]{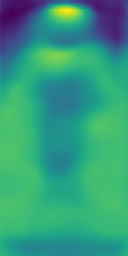} & 
    \includegraphics[width=0.04\textwidth]{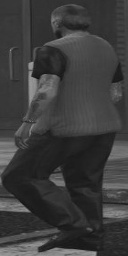} & 
    \includegraphics[width=0.04\textwidth]{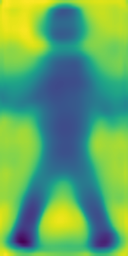}
    \\
    & \multicolumn{5}{c|}{Celeb-reID} & \multicolumn{5}{c||}{VC-Clothes} & \multicolumn{4}{c|}{Celeb-reID} & \multicolumn{4}{c}{VC-Clothes} \\
    & \multicolumn{10}{c||}{(a)} & \multicolumn{8}{c}{(b)}
    \end{tabular}
    \caption{Examples of the samples generated by the proposed feature augmentation module using Celeb-reID~\cite{Huang_2020_TCSVT} and VC-Clothes~\cite{wan2020person} datasets. The reconstructed images are augmented using the id-relevant features of persons in each image of the first row and the id-irrelevant features of persons in the first column. (a) Reconstructed images with negative samples. (b) Reconstructed images with positive samples. The augmented examples are visualized using the Viridis color map. \vspace{-0.2cm}}  \label{fig:mix}
\end{figure*}

The results of this study on Celeb-reID and Celeb-reID-light datasets are tabulated in \tref{tab:ablation_loss}. The re-identification accuracy improves by $8.0\%$~(Rank1) / $4.5\%$~(mAP) on Celeb-reID and $-0.6\%$~(Rank1) / $0.5\%$~(mAP) on Celeb-reID-light for the combination of $\mathcal{L}_{cls}+\mathcal{L}_{tri}+\mathcal{L}_{sim}$. The accuracy improves by $13.1\%$~(Rank1) / $21.4\%$~(mAP) on Celeb-reID and $6.2\%$~(Rank1) / $2.7\%$~(mAP) on Celeb-reID-light for the combination of $\mathcal{L}_{cls}+\mathcal{L}_{tri}+\mathcal{L}_{rec}$. Also, the accuracy increases by $15.2\%$~(Rank1) / $25\%$~(mAP) on Celeb-reID and $14.3\%$~(Rank1) / $7.7\%$~(mAP) on Celeb-reID-light for the loss function with the proposed negative feature augmentation, $\mathcal{L}_{cls}+\mathcal{L}_{tri}+\mathcal{L}_{aug}^{n}$, and $15.8\%$~(Rank1) / $22.3\%$~(mAP) on Celeb-reID and $8.8\%$~(Rank1) / $4.9\%$~(mAP) on Celeb-reID-light for the loss function with the positive feature augmentation, $\mathcal{L}_{cls}+\mathcal{L}_{tri}+\mathcal{L}_{aug}^{p}$.    
This analysis verifies that each proposed module contributes to performance enhancement. Besides, by combining the proposed modules together, the accuracy significantly improves by $17.9\%$~(Rank1) / $26.8\%$~(mAP) on Celeb-reID and $16.9\%$~(Rank1) / $9.9\%$~(mAP) on Celeb-reID-light, as expected. This demonstrates that the proposed framework is effective to improve re-identification performance because of feature representation by distribution, and as a result, the framework is able to generate difficult samples during training.  

\begin{table*}
    \centering
    \caption{Comparisons of generalization performance. All methods are trained using the Celeb-reID-light and tested on the Celeb-reID, VC-Clothes, and LTCC~(clothes-changing) datasets.}
    \label{tab:generalization_capability}
    {\renewcommand{\arraystretch}{1.2}
    \begin{tabular}{l|ccc|ccc|ccc}
        \hline
        
        \multirow{2}{*}{Method} & 
        \multicolumn{3}{c|}{Celeb-reID-light $\rightarrow$ Celeb-reID} &
        \multicolumn{3}{c|}{Celeb-reID-light $\rightarrow$ VC-Clothes} &
        \multicolumn{3}{c}{Celeb-reID-light $\rightarrow$ LTCC}\\ \cline{2-10}
        & Rank1 & Rank5 & mAP & Rank1 & Rank5 & mAP & Rank1 & Rank5 & mAP \\ \hline \hline
        ReIDCaps & 43.4 & 56.9 & 6.9 & 16.0 & 29.7 & 6.4 & 5.9 & 13.0 & 2.5\\
        LaST & 43.7 & 57.0 & 6.3 & 6.5 & 14.4 & 3.7 & 3.3 & 9.2 & 2.4 \\
        RCSANet  & \textbf{50.4} & - & 11.1 & - & - & - & - & - & - \\ \hline
        \textbf{Ours} & 47.3 & \textbf{63.5} & \textbf{11.9} & \textbf{35.6} & \textbf{45.0} & \textbf{18.8} & \textbf{8.2} & \textbf{17.6} & \textbf{4.1} \\
        \hline
    \end{tabular}} 
\end{table*}

\begin{figure}[!t]
    \centering\
    \begin{tabular}{@{}c@{~}c@{~}c@{~}|c@{~}c@{~}c@{~}|c@{~}c@{~}c@{}}
    \includegraphics[width=0.045\textwidth]{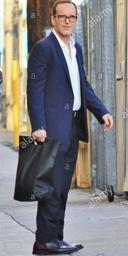} & 
    \includegraphics[width=0.045\textwidth]{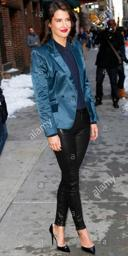} & 
    \includegraphics[width=0.045\textwidth]{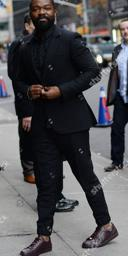} &
    \includegraphics[width=0.045\textwidth]{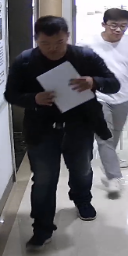} & 
    \includegraphics[width=0.045\textwidth]{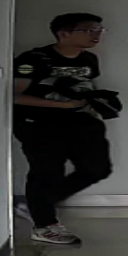} & 
    \includegraphics[width=0.045\textwidth]{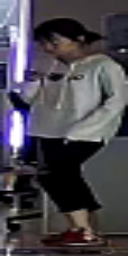} & 
    \includegraphics[width=0.045\textwidth]{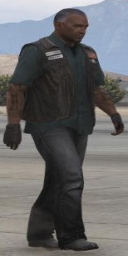} & 
    \includegraphics[width=0.045\textwidth]{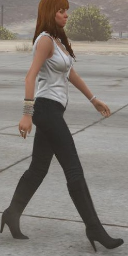} &
    \includegraphics[width=0.045\textwidth]{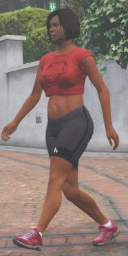}
    \\
    \includegraphics[width=0.045\textwidth]{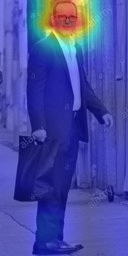} & 
    \includegraphics[width=0.045\textwidth]{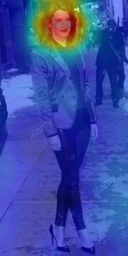} & 
    \includegraphics[width=0.045\textwidth]{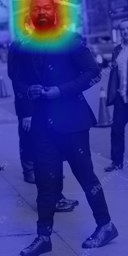} &
    \includegraphics[width=0.045\textwidth]{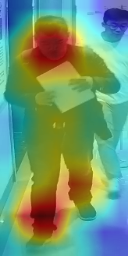} & 
    \includegraphics[width=0.045\textwidth]{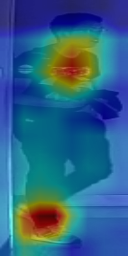} & 
    \includegraphics[width=0.045\textwidth]{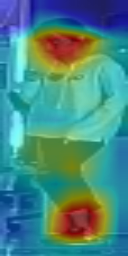} & 
    \includegraphics[width=0.045\textwidth]{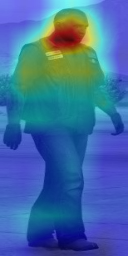} & 
    \includegraphics[width=0.045\textwidth]{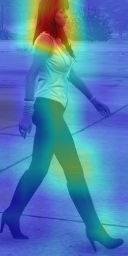} & 
    \includegraphics[width=0.045\textwidth]{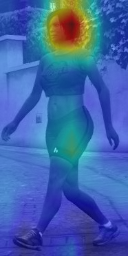}
    \\ 
    \multicolumn{3}{c|}{Celeb-reID} &  \multicolumn{3}{c|}{LTCC} &  \multicolumn{3}{c}{VC-Clothes}
    \end{tabular}
    \caption{Examples of the feature activation mapping $E_{g}$ using Celeb-reID~\cite{Huang_2020_TCSVT}, LTCC~\cite{qian2020long}, and VC-Clothes~\cite{wan2020person} dataset. The feature activation maps highlight the identity-specific discriminative regions.} \vspace{-0.3cm} \label{fig:activation}
\end{figure}

In addition to the performance comparison, some examples of retrieval results with different combinations of loss functions are shown in \fref{fig:ranking}. The retrieval results of the proposed framework are much more insensitive to cloth changing compared to the baseline~(``$\mathcal{L}_{cls}+\mathcal{L}_{tri}$'') on both Celeb-reID and LTCC datasets. In addition, the proposed framework can accurately identify the same person as the most similar one in most example cases. For example, the proposed framework using $\mathcal{L}_{sim}$ and $\mathcal{L}_{rec}$ can identify the same person among other identities wearing similar clothes, as shown in the first and second examples of Celeb-reID. For the fourth example of Celeb-reID, the proposed framework can consistently identify the same person under cloth changing. For the LTCC dataset, the model using $\mathcal{L}_{cls}$ and $\mathcal{L}_{tri}$ fails to identify the same person as a query in most cases. On the contrary, the proposed framework can identify the same person regardless of various pose changes. 

\subsubsection{Sampling Independent Feature Representation}
To visualize the role of the sample independent maximum discrepancy loss $\mathcal{L}_{sim}$ in \eref{eq:L_sim}, we conduct a t-distributed stochastic neighbor embedding~(t-SNE)~\cite{van2008visualizing} analysis on the proposed framework using Celeb-reID and LTCC datasets. For this study, we randomly select $11$ identities for each identity. The structural changes of the selected samples are then plotted every $5$-epoch in Celeb-reID and every $50$-epoch in LTCC dataset, applying/not applying the sample independent maximum discrepancy loss in \eref{eq:L_sim}. The effectiveness of the proposed all sample-based discriminative learning is intuitively verified by the distribution perspective in the feature embedding space in \fref{fig:tsne_cluster}. The model with the loss $\mathcal{L}_{sim}$ over all samples performs better clustering than the model without it, as proved in \fref{fig:tsne_cluster}. 

For the Celeb-reID dataset, the clusters of persons $3$, $9$, and $10$ are not separable in the third figure of the first row in \fref{fig:tsne_cluster}, and persons $0$ and $8$ can be split into $2$-group in the third figure of the second row. By contrast, the points of persons $3$, $9$, and $10$ are closely aligned to a cluster in the first figure of the first row in \fref{fig:tsne_cluster}. The clusters of persons $0$ and $8$ are clearly separable in the third figure of the second row. Furthermore, the plots indicate that the model converges faster, as the structure of samples is already stable in the first figure of the second row. This analysis is consistent with the LTCC dataset. The points of person $8$ are not gathered to one cluster in the last figure of the first row, while the points are much pulled together in the last figure of the second row. The clusters of persons $3$ and $10$ are inseparable in the figures of the first row. With the sample independent maximum discrepancy loss, the clusters of those people are divisible in the last figure of the second row.  

\subsubsection{Learnable Feature Augmentation}
In \fref{fig:mix}, we visualize the reconstructed images using the identity-relevant features of persons in the first column and the identity-irrelevant features of persons in the first row on Celeb-reID and VC-Clothes datasets. As observed from the examples, id-irrelevant features~(\eg clothes, background, etc.) vary across row-wise images while the features are visible with insignificant changes across column-wise images. This infers that the proposed framework explicitly separates id-relevant features and id-irrelevant features without the supervision of appearance changes.    

The examples of the feature activation map $E_{g}$ are also visualized to analyze where the model pays more attention to maximizing identity discrimination.  In \fref{fig:activation}, we highlight the identity-specific discriminative regions learned by the proposed pseudo-ground-truth generation module. Interestingly, id-relevant features locate around face parts in most samples, and id-irrelevant features are likely to be represented by background or other body parts. This suggests that our approach to exploring the most id-relevant part is as expected, and thus the proposed hard sample generation helps to uncouple feature embeddings.  

\subsubsection{Generalization Performance} We evaluate the generalization performance of the proposed framework against ReIDCaps, LaST, and RCSANet. In \tref{tab:generalization_capability}, all methods are trained using the Celeb-reID-light dataset and tested on the Celeb-reID, VC-Clothes, and LTCC datasets. The results demonstrate that the proposed framework has better generalization performance even for the unseen VC-Clothes and LTCC datasets than the compared methods.

\section{Conclusion} \label{sec:conclusion}
In this work, we proposed a novel framework that is robust against sample selection for long-term re-identification. The key contribution of this work is to model each sample as a cluster and extract disentangled features guided by it,  guaranteeing the model's performance independent of sample selection strategies.  With the learned feature distribution, the proposed framework could generate hard samples that improved the model's discriminability. Extensive experiments were performed to validate the effectiveness of the proposed framework. The proposed framework outperformed prior state-of-the-art models on common benchmark datasets.
\vspace{0.3cm}

\bibliographystyle{IEEEtran}
\bibliography{refs}

\vfill

\end{document}